\documentclass[10pt]{article}

\usepackage{bm}

\usepackage{fullpage}  

\usepackage{url}
\usepackage[ruled,linesnumbered]{algorithm2e}

\usepackage{multirow} 
\usepackage{hhline}  
\usepackage{verbatim}
\usepackage{graphicx}
\usepackage{amsmath,amssymb,amsfonts,graphicx,amsthm,mathtools,nicefrac}
\usepackage{lscape}
\usepackage{color}
\usepackage{authblk}
\usepackage{enumerate}
\usepackage{subcaption}
\usepackage{booktabs}

\usepackage{pifont}       
\usepackage{bbding}       
\usepackage{fontawesome}  
 
\newcommand{\cmark}{\ding{51}}
\newcommand{\xmark}{\ding{55}}

\allowdisplaybreaks
\newcommand\numberthis{\addtocounter{equation}{1}\tag{\theequation}}
\newtheorem{definition}{Definition}[section]
\newtheorem{theorem}{Theorem}[section]
\newtheorem{assumption}{Assumption}[section]
\newtheorem{remark}{Remark}[section]

\numberwithin{equation}{section}

\newtheorem{lemma}[theorem]{Lemma}

\DeclareMathOperator{\Unif}{Unif}

\DeclareMathOperator{\bias}{bias}

\def\la {\left\langle}
\def\ra {\right\rangle} 
\def \lb{\left(}
\def \rb{\right)}

\newcommand{\matsnorm}[2]{\left\| #1\right\|_{{#2}}}

\newcommand{\opnorm}[1]{\ensuremath{\matsnorm{#1}{}}}

\newcommand{\twonorm}[1]{\ensuremath{\matsnorm{#1}{\footnotesize{2}}}}

\newcommand{\bfm}[1]{\bm{#1}}

\renewcommand{\Pr}[2][]{\mathbb{P}_{#1} \left\{ #2 \rule{0mm}{3mm}\right\}}

\newcommand{\E}[2][]{\mathbb{E}_{#1} \left\{ #2 \rule{0mm}{3mm}\right\}}

\def\va{\bfm a}     
\def\vb{\bfm b}

  \def\mF{\bfm F}  
\def\vg{\bfm g}     
   \def\mH{\bfm H}  
   \def\mI{\bfm I}

   \def\mM{\bfm M}

     \def\R{\mathbb{R}}

\def\vu{\bfm u}     
\def\vv{\bfm v}     
\def\vw{\bfm w}     
\def\vx{\bfm x}     
     
\def\vz{\bfm z}     

\def \vtheta {\bfm \theta}

\def\calA{{\cal  A}}

\def\calG{{\cal  G}} 
\def\calH{{\cal  H}}

\def\calM{{\cal  M}} 
\def\calN{{\cal  N}} 
\def\calO{{\cal  O}}

\def\calS{{\cal  S}}

\newcommand{\bfsym}[1]{\bm{#1}}

             \def\hbtheta {\hat{\bfsym {\theta}}}

\def \tran {\mathsf{T}}

\def \bzero{\bm 0}

\usepackage{bbm}

\def \hw{\widehat{\vw}}
\def \td {\tilde{d}}

\DeclareMathOperator*{\argmin}{arg\,min}

\begin{document}
	
\title{Global Convergence of Natural Policy Gradient with Hessian-aided Momentum Variance Reduction}
\author[1]{Jie Feng}
\author[1]{Ke Wei}
\author[2]{Jinchi Chen}
\affil[1]{School of Data Science, Fudan University, Shanghai, China.}
\affil[2]{School of Mathematics, East China University of Science and Technology, Shanghai, China.\vspace{.15cm}}
\date{\today}

\maketitle
\begin{abstract}
Natural policy gradient (NPG) and its variants are  widely-used policy search methods in reinforcement learning. Inspired by prior work,  
a new NPG variant coined NPG-HM is developed in this paper, which utilizes the Hessian-aided momentum technique for variance reduction, while the sub-problem is solved via the stochastic gradient descent method.
It is shown that NPG-HM can achieve the global last iterate  $\varepsilon$-optimality with a sample complexity of $\calO(\varepsilon^{-2})$, which is the best known result for natural policy gradient type methods under the generic Fisher non-degenerate policy parameterizations. The convergence analysis is built upon a relaxed weak gradient dominance property tailored for NPG under the compatible function approximation framework, as well as a neat way to decompose the error  when handling the sub-problem. Moreover, numerical experiments on Mujoco-based environments demonstrate the superior performance of NPG-HM over other state-of-the-art policy gradient  methods.
\end{abstract}

\section{Introduction}

{Reinforcement Learning (RL), which attempts to maximize long-term reward  in a sequential decision-making task, has found  a wide range of applications, for example in robotics~\cite{kober2013reinforcement}, game playing~\cite{garisto2019google}, and recommendation systems~\cite{afsar2022reinforcement}}. A standard RL setting can be represented as a Markov Decision Process (MDP),  {denoted} $\calM=(\calS,\calA,P, r,\gamma)$, where $\calS$ is the state space, $\calA$ is the action space, $P(s'|s,a)$ denotes the probability of transitioning into state $s'$ after taking action $a$ at state $s$, $r(s,a,s')$ is the immediate reward, and $\gamma\in[0,1)$ is the discount factor. Assume the agent takes actions following a policy that is parameterized by a vector $\vtheta\in\R^d$, denoted $\pi_{\vtheta} :\calS\to\Delta(\calA)$. That is, at any time step $h$, the agent takes an action $a^h\sim\pi_{\vtheta}(\cdot|s^h)$, and then transitions to the next state $s^{h+1}\sim P(\cdot|s^h,a^h)$. In this paper we focus on the infinite horizon discounted setting. Given a trajectory $\tau = (s^{0}, a^{0}, r^{0},s^1,a^1,r^1,\dots)$ induced by a policy $\pi_{\vtheta}$, where $a^{h}\sim \pi_{\vtheta}(\cdot|s^{h})$, $r^h = r(s^h,a^h,s^{h+1})$, and $s^{h+1}\sim P(\cdot|s^h,a^h)$, the state value function $V^{\pi_{\vtheta}}: \calS \to \mathbb{R}$ is defined as the average discounted sum of immediate rewards,
\begin{align*}
V^{\pi_{\vtheta}}(s) = 
\E[\tau]{\sum_{h=0}^{\infty} \gamma^h  r^h | s^0 = s}.
\numberthis\label{eq:value func}
\end{align*}
Under an initial state distribution $\rho$, the goal of RL is to find a parameter $\vtheta$ that maximizes the expected cumulative rewards:
\begin{align*}
\max_{\vtheta\in\R^d}  \;  J_\rho(\vtheta) := \E[s\sim\rho]{V^{\pi_{\vtheta}}(s)}=\E[\tau\sim p_{\vtheta}]{\sum_{h=0}^{\infty} \gamma^h r^h }, 
\end{align*}
where $p_{\vtheta}$ denotes the  trajectory distribution induced by $\pi_{\vtheta}$:
\begin{align*} \numberthis\label{dist of tau}
p_{\vtheta}(\tau)= \rho(s^0) \prod_{h=0}^{\infty} \pi_{\vtheta} (a^h|s^h) P(s^{h+1}|s^{h},a^{h}).
\end{align*} 

Once RL is formulated as an optimization problem over the parameter space, a variety of optimization methods are available for seeking the optimal policy parameter, including  vanilla policy gradient (PG), natural policy gradient (NPG)~\cite{kakade2001natural,bagnell2003covariant}, trust region policy optimization (TRPO)~\cite{schulman2015trust}, proximal policy optimization (PPO)~\cite{schulman2017proximal}, actor-critics methods~\cite{konda1999actor, haarnoja2018soft, peters2008natural}, and policy mirror ascent (PMA) \cite{tomar2020mirror, lan2023policy}. Compared with valued-based methods, policy optimization   can be readily extended to high dimensional discrete or continuous action spaces via different parameterized policies, and is also amenable to a detailed analysis. Indeed, the convergence analyses of various policy optimization methods have received a lot of attention recently. We will first give a brief discussion towards this line of research. 


\subsection{Related works}
\paragraph{Convergence of exact policy gradient methods.}
Recently, a series of works have studied the global optimality (i.e., the convergence of $J_\rho(\vtheta_t)$ to $J_\rho^*$, where $\vtheta_t$ is the output of an algorithm and $J_\rho^*$ is the optimal objective) of policy gradient methods when the policy gradient is exactly evaluated. 
Regarding the simplex parameterization, the sublinear convergence of the corresponding projected policy gradient (PPG) method with a constant step size has been investigated in \cite{agarwal2021theory,Zhang_Koppel_Bedi_Szepesvari_Wang_2020,xiao2022convergence,liu2023projected}. Furthermore, it can be  shown  that PPG is able to achieve the exact convergence in a finite number of iterations \cite{liu2023projected}. The sublinear convergence of the constant step size PG under the softmax parameterization (softmax PG) is established in \cite{mei2020global,mei2021leveraging}, where the linear convergence is further proved for the entropy-regularized softmax PG. Softmax NPG, a preconditioned version of softmax PG, also enjoys the sublinear convergence for a constant step size \cite{agarwal2021theory}, while its local linear convergence is established in \cite{khodadadian2021linear}. Moreover, a linear convergence rate for entropy-regularized softmax NPG is obtained in \cite{cen2022fast}. Noticing that softmax NPG in the policy space can be viewed as a special case of the policy mirror ascent  method, the sublinear convergence  can also be extended to the general PMA method with a constant step size \cite{xiao2022convergence}. In addition, the linear convergence of PMA has been established for  geometrically increasing step sizes in the same work. 
Moreover, the convergence of PMA with different regularizers has been studied in  \cite{lan2023policy, zhan2021policy, li2022homotopic}.

\paragraph{Sample complexity for first-order stationary point convergence.}
In practice, the policy gradient cannot be computed exactly, but should be estimated from random samples. Thus a line of research is devoted to the first-order stationary point analysis of policy optimization methods using sampled trajectories under general policy parameterizations, which follows closely the works in stochastic optimization. The vanilla PG is proved to require $\calO(\varepsilon^{-4})$ random trajectories to reach an $\varepsilon$-stationary point such that $\|\nabla J_\rho(\vtheta_t)\|_2 \le \varepsilon$~\cite{yuan2022general,liu2020improved,zhang2020global, qiong2021non,papini2021safe}. Since the simple Monte Carlo estimate of policy gradient suffers from high variance, many variance reduction techniques have been introduced into PG to improve the sample efficiency and accelerate the algorithm, see for example \cite{xu2019sample, yuan2020stochastic, zhang2021convergence, gargiani2022page, huang2020momentum, salehkaleybar2022adaptive} and references therein. After the introduction of variance reduction, the sample complexity can be reduced from $\calO(\varepsilon^{-4})$ to $\calO(\varepsilon^{-3})$, which is overall optimal. 


\paragraph{Sample complexity for global convergence.} 
There has  been a surge of research studying the sample complexity of the  sample-based policy gradient methods for the global convergence, which is also the focus of this paper. The property of gradient dominance, also known as Polyak-Lojasiewicz (PL) condition, is often used in this line of research. In a nutshell, gradient dominance 
prevents the gradient from vanishing unless reaching the global optimum. The sample-based NPG and Q-NPG methods under the log-linear policy have been studied within a compatible function approximation framework \cite{agarwal2021theory}, establishing the $\calO(\varepsilon^{-6})$ sample complexity through a connection between the global optimum and the update direction. This result has been improved to $\calO(\varepsilon^{-2})$ for the geometrically increasing step sizes  by viewing the sample-based NPG and Q-NPG methods as the approximate policy mirror optimization (APMO) methods \cite{yuan2022linear}. Indeed,  NPG performs intrinsically more and more akin to policy iteration with large enough learning rates, implying the geometric convergence. Moreover, the global convergence of APMO  has been studied recently in \cite{alfano2023novel} for the general Bregman projected policy and an $\calO(\varepsilon^{-4})$ sample complexity is obtained for the special case of shallow neural network parameterization. For the general Fisher-non-degenerate policy, a relaxed weak gradient domination property is introduced in \cite{ding2021global}, and it is shown that momentum-based PG requires an $\calO(\varepsilon^{-3})$ sample complexity to achieve the global average-regret convergence, defined as
$$
    J_\rho^* - \frac{1}{T}\sum^{T-1}_{t=0}\E{J_\rho(\vtheta_t)}\le\varepsilon +\calO(\sqrt{\varepsilon_\mathrm{bias}}),
$$
where $\varepsilon_\mathrm{bias}$ is a parameter introduced in the compatible function approximation framework. Furthermore, the $\calO(\varepsilon^{-3})$ sample complexity has been established in \cite{yuan2022general} for sample-based vanilla PG to achieve the global optimality of the form $\min_t \{J_\rho^*-\E{J_\rho(\vtheta_t)}\}\leq \varepsilon +\calO(\sqrt{\varepsilon_\mathrm{bias}})$.
In contrast, an (N)-HARPG method is proposed in \cite{fatkhullin2023stochastic} which only requires an $\calO(\varepsilon^{-2})$ sample complexity to achieve  the global last-iterate convergence, expressed as 
$$
    J_\rho^* - \E{J(\vtheta_T)} \le \varepsilon+\calO(\sqrt{\varepsilon_\mathrm{bias}}).
$$
An innovation in \cite{fatkhullin2023stochastic} is  the Hessian-aided scheme for computing the gradient difference when constructing the moment-based variance reduction estimator, which can avoid the unverifiable assumption in the analysis when using importance sampling to construct the estimator. 
In addition, a double-loop variance-reduced NPG with batch gradients is proposed in  \cite{liu2020improved} and  the $\calO(\varepsilon^{-3})$ sample complexity for the global average-regret convergence has been established.  This sample complexity has been improved to $\calO(\varepsilon^{-2})$ for accelerated natural policy gradient (ANPG) \cite{mondal2023improved}, in which an accelerated gradient descent procedure is utilized  to solve the sub-problem concerning the NPG update direction.

\subsection{Main contributions}
The main contributions of this paper are summarized as follows:

\begin{itemize}
    \item Inspired by \cite{fatkhullin2023stochastic}, a sample-based natural policy gradient method with Hessian-aided momentum variance reduction (NPG-HM) is proposed. 
    Compared with the variance reduction NPG method developed in \cite{liu2020improved}, NPG-HM is a single-loop algorithm that avoids the usage of importance sampling when constructing the unbiased policy gradient estimator.  Numerical experiments on Mujoco-based environments demonstrate that
NPG-HM outperforms the other state-of-the-art policy gradient  methods.
    
    \item The global convergence of NPG-HM is studied when the sub-problem for computing the NPG update direction is solved via the stochastic gradient descent method. It is shown that it requires an $\calO(\varepsilon^{-2})$ sample complexity for NPG-HM to achieve  the global last-iterate $\varepsilon$-optimality under the general Fisher-non-degenerate policy, differing from the  average-regret convergence results established in \cite{liu2020improved} and \cite{mondal2023improved} for sample-based NPG methods. Though our analysis is also inspired by \cite{fatkhullin2023stochastic}, the extension is by no means trivial since it involves solving a sub-problem through SGD in NPG-HM. More precisely, an innovative way to decompose the error is used when
handling the sub-problem, and then a recursive sequence is developed based on the decomposition,  see Lemma~\ref{lemma ascent 2} and Lemma~\ref{lemma 3}, respectively. A comparison with other sample-based PG and NPG methods which also enjoy  global convergence is presented in Table~\ref{tab: comp}. 
\end{itemize}
\begin{table}[ht!]
\centering
\caption{Sample complexities of sample-based PG and NPG methods to achieve a global $\varepsilon$-optimality. 
``{\bf IS-free}'' means  the importance sampling technique is not required, 
while ``{\bf Last-iterate}'' indicates the global last-iterate convergence.
}\label{tab: comp}
\renewcommand{\arraystretch}{1.2}
\begin{tabular}{ c|c|c|c|c} 
\toprule[1.2pt]
 Algorithm & Parameterization &Sample Complexity & IS-free & Last-iterate \\
\hline
NPG \cite{agarwal2021theory} & Log-linear & $\varepsilon^{-6}$ & \cmark & \xmark \\ 
\hline
NPG \cite{yuan2022linear} & Log-linear & $\varepsilon^{-2}$ & \cmark & \cmark \\
\hline
PG ~\cite{yuan2022general} & Fisher-non-degenerate & $\varepsilon^{-3}$ & \cmark &\xmark\\
\hline
STORM-PG ~\cite{ding2021global} & Fisher-non-degenerate & $\varepsilon^{-3}$ & \xmark &\xmark\\
\hline
(N)-HARPG~\cite{fatkhullin2023stochastic}&Fisher-non-degenerate & $\varepsilon^{-2}$ & \cmark &  \cmark\\
\hline
NPG-SRVR~\cite{liu2020improved}& Fisher-non-degenerate & $\varepsilon^{-2.5}+\varepsilon^{-3}$ &\xmark & \xmark\\
\hline
ANPG ~\cite{mondal2023improved}& Fisher-non-degenerate& $\varepsilon^{-2}$ &\cmark & \xmark \\
\hline
\textbf{NPG-HM} & Fisher-non-degenerate &$\varepsilon^{-2}$ &\cmark & \cmark\\
\toprule[1.2pt]
\end{tabular}
\end{table}

\subsection{Paper organization}
The rest of this paper is organized as follows. In Section~\ref{sec: Preliminaries}, we review some results and methods about RL and variance reduction. The description of the proposed NPG-HM and the main theoretical result is presented in Section~\ref{sec:alg and res}. The proof of the main result is provided in Section~\ref{sec: proof of thm}.  Numerical experiments conducted to compare NPG-HM with other policy gradient methods can be found in Section~\ref{sec: numerical experiments}. In Section~\ref{sec: conclusions}, we conclude this paper with a few interesting future research directions. Finally, Section~\ref{sec: proofs of lmm} is devoted to the proofs of some technical lemmas.

\section{Preliminaries} \label{sec: Preliminaries}
Recalling the definition of state value function in~\eqref{eq:value func}, there are another two value functions which are closely related.  Given a state-action pair $(s,a)$, define the state-action value function (or Q-function) $Q^{\pi_{\vtheta}}: \calS \times \calA \to \mathbb{R}$ as follows:
\begin{align*}
	Q^{\pi_{\vtheta}}(s,a) = \E[\tau]{\sum_{h=0}^{\infty} \gamma^h  r^h | s^0 = s,a^0=a}.
\end{align*}
It is easy to show that 
\begin{align*}
V^{\pi_{\vtheta}}(s)=\E[a\sim\pi_{\vtheta}(\cdot|s)]{Q^{\pi_{\vtheta}}(s,a)}.
\end{align*}
In addition, the advantage function $A^{\pi_{\vtheta}}: \calS \times \calA \to \mathbb{R}$ which measures how well a single action $a$ behaves compared with the average value is defined as 
\begin{align*}
	A^{\pi_{\vtheta}}(s,a) = Q^{\pi_{\vtheta}}(s,a)-V^{\pi_{\vtheta}}(s).
\end{align*}

\subsection{Policy gradient}
Define the state visitation distribution $d_{\rho,\pi_{\vtheta}}$ induced by policy $\pi_{\vtheta}$ as
\begin{align*}
d_{\rho,\pi_{\vtheta}}(s)= (1-\gamma)\E[s^0\sim \rho]{\sum_{h=0}^{\infty} \gamma^h\Pr{s^h = s |s^0,\pi_{\vtheta}}},
\end{align*}
where $\Pr{s^h = s |s^0,\pi_{\vtheta}}$ is the probability that the state $s$ is visited at time step $h$, given that the trajectory starts from $s^0$ and follows policy $\pi_{\vtheta}$. Moreover, define the state-action visitation distribution $\td_{\rho, \pi_{\vtheta}}(s,a)$  as
\begin{align*}
\numberthis\label{state action dist}
    \td_{\rho, \pi_{\vtheta}}(s,a) = d_{\rho,\pi_{\vtheta}}(s)\cdot \pi_{\vtheta}(a|s).
\end{align*}
Then the policy gradient of $J_\rho(\vtheta)$ with respect to $\vtheta$ can be expressed as (\cite{sutton2018reinforcement})
\begin{align*}
    \nabla J_\rho(\vtheta) &= \frac{1}{1-\gamma}\E[(s,a)\sim \td_{\rho, \pi_{\vtheta}}]{\nabla \log \pi_{\vtheta}(a|s) Q^{\pi_{\vtheta}}(s,a)} \\
    &=\frac{1}{1-\gamma}\E[(s,a)\sim \td_{\rho, \pi_{\vtheta}}]{\nabla \log \pi_{\vtheta}(a|s) A^{\pi_{\vtheta}}(s,a)},
\end{align*}
where the second equality relies on the assumption $\sum_{a\in\mathcal{A}}\pi_{\vtheta}(a|s)=1, \forall\,  \vtheta\in\R^d$ that will be made throughout this paper.
In terms of the expectation with respect to the trajectory, $\nabla J_\rho(\vtheta)$ has the following alternative expression:
\begin{align*}
    \nabla J_\rho(\vtheta)  = \E[\tau\sim p_{\vtheta}]{\sum_{h=0}^{\infty}\lb \sum _{i=h}^{\infty} \gamma^{i} r^{i}\rb \nabla \log \pi_{\vtheta}(a^{h}|s^{h})},\numberthis\label{gradient}
\end{align*}
where $p_{\vtheta}$ is given in \eqref{dist of tau}. In addition, the policy Hessian $\nabla^2 J(\vtheta)$ which will be used later is given by (\cite{shen2019hessian})
\begin{align*}
    \nabla^2 J_\rho(\vtheta) = \E[\tau\sim p_{\vtheta}]{\nabla \phi(\tau; \vtheta) (\nabla \log p_{\vtheta}(\tau))^\tran + \nabla^2 \phi(\tau; \vtheta)},\numberthis\label{hessian}
\end{align*}
where $\phi(\tau;{\vtheta})=\sum_{h=0}^{\infty}\lb \sum _{i=h}^{\infty} \gamma^{i} r^{i}\rb\log \pi_{\vtheta}(a^{h}|s^{h})$.  

Given the expression of policy gradient, the vanilla policy gradient method simply updates the policy parameter along the gradient ascent direction:
\begin{align*}
    \vtheta_{t+1} = \vtheta_t + \alpha_t \nabla  J_\rho(\vtheta_t),
\end{align*}
where $\alpha_t$ is the learning rate. As previously mentioned, the policy gradient needs to be estimated from random samples in practice. The expression in \eqref{gradient} provides a natural unbiased estimator based on  random trajectories. However, the length of sampled trajectories cannot be infinite in practice. Thus we instead consider the following truncated estimator:
\begin{align*}
\vg(\tau; \vtheta) := \sum_{h=0}^{H-1}\lb \sum _{i=h}^{H-1} \gamma^{i} r^{i}\rb \nabla \log \pi_{\vtheta}(a^{h}|s^{h}).\numberthis\label{eq:grad1}
\end{align*}
Note that $\vg(\tau; \vtheta)$ is a biased estimator for $\nabla J_\rho(\vtheta)$, but  an unbiased estimator for $\nabla J_\rho^H(\vtheta)$, where 
\begin{align*}
J_\rho^H (\vtheta) =\E[\tau\sim p_{\vtheta}]{\sum_{h=0}^{H-1} \gamma^h r^h },
\end{align*}
with \begin{align}
\label{truncated tau}
p_{\vtheta}(\tau)= \rho(s^0) \prod_{h=0}^{H-1} \pi_{\vtheta} (a^h|s^h) P(s^{h+1}|s^{h},a^{h}).
\end{align}
Similarly, we will consider the following truncated and biased estimator for $\nabla^2 J(\vtheta)$:
\begin{align*}
    \mH(\tau; \vtheta) := \nabla \phi(\tau; \vtheta) (\nabla \log p_{\vtheta}(\tau))^\tran + \nabla^2 \phi(\tau; \vtheta),\numberthis\label{eq:hess1}
\end{align*}
where $\phi(\tau;{\vtheta})=\sum_{h=0}^{H-1}\lb \sum _{i=h}^{H-1} \gamma^{i} r^{i}\rb\log \pi_{\vtheta}(a^{h}|s^{h})$, which again is  an unbiased estimator for $\nabla^2 J^H_\rho(\vtheta)$.

\subsection{Natural policy gradient}
As an important variant of PG, NPG utilizes the Fisher Information Matrix (FIM) as a preconditioner to refine the gradient direction based on the underlying structure of the parameterized policy space, which can be approximately viewed as a second-order method. 
The FIM $\mF(\vtheta)$ is defined by
\begin{align*}\numberthis\label{eq: fim def}
\mF(\vtheta) = \E[(s,a)\sim \td_{\rho, \pi_{\vtheta}}]{\nabla \log \pi_{\vtheta}(a|s) \left( \nabla \log \pi_{\vtheta}(a|s)\right)^\tran},
\end{align*}
where $\td_{\rho, \pi_{\vtheta}}$ represents the state-action visitation distribution defined in \eqref{state action dist}. Given the FIM, the NPG update takes the following form:
\begin{align*}
	\numberthis\label{eq:npg}
	\vtheta_{t+1} = \vtheta_t + \alpha_t \mF^{\dagger}(\vtheta_t) \nabla J_\rho(\vtheta_t),
\end{align*}
where $\alpha_t$ is the learning rate, 
and $\mF^{\dagger}(\vtheta)$ is the Moore-Penrose pseudoinverse of $\mF(\vtheta)$.

Moreover, for any $\vtheta$ and $\vw$, define the compatible function approximation error  as follows:
\begin{align}
\label{compatible function}
L(\vw ; \vtheta)=\frac{1}{2}\E[(s,a)\sim \td_{\rho, \pi_{\vtheta}}]{ \left((1-\gamma) \vw^{\top} \nabla\log \pi_{\vtheta}(a | s) - A^{\pi_{\vtheta}}(s, a) \right)^2}.
\end{align}
Then it is easy to see that $\vw_{t}^\ast=\mF^{\dagger}(\vtheta_t) \nabla J_\rho(\vtheta_t)$ is a minimizer of \eqref{compatible function}. Thus, the NPG update direction 
can be obtained by solving the following minimization problem:
\begin{align*}
\vw_{t}^\ast = \argmin_{\vw} L(\vw ; \vtheta_t).
\end{align*}
It is also evident that 
$\vw_t^\ast$ is a minimizer of 
\begin{align}
\label{sub-prob of npg}
    \min_{\vw\in\R^d} \left\{  \frac{1}{2} \E[(s,a)\sim \td_{\rho, \pi_{\vtheta_t}} ]{ \left(\vw^{\top} \nabla\log \pi_{\vtheta_t}(a | s) \right)^2 }- \vw^\tran\nabla J_\rho(\vtheta_t) \right\}.
\end{align}



\subsection{Hessian-aided momentum variance reduction}

Recall that the momentum-based gradient estimator from~\cite{ding2021global, huang2020momentum} is given by
\begin{align*}\numberthis\label{eq:momentum}
	\vv_t = \beta_t \vg(\tau_t; \vtheta_t) + (1-\beta_t)  \left( \vv_{t-1} + \vg(\tau_t;\vtheta_t) - \omega(\tau_t | \vtheta_{t-1}, \vtheta_t)\cdot \vg(\tau_t;\vtheta_{t-1}) \right),
\end{align*}
where $\beta_t\in(0,1]$ is the momentum coefficient, and  $\omega(\tau_t |\vtheta_{t-1}, \vtheta_t)$ is the importance sampling weight:
\begin{align}
	\label{eq important sampling}
	\omega(\tau_t |\vtheta_{t-1},\vtheta_t)= \frac{p_{\vtheta_{t-1}}(\tau_t) }{ p_{\vtheta_t}(\tau_t)} = \prod_{h=0}^{H-1}\frac{\pi_{\vtheta_{t-1}}(a^h|s^h)}{\pi_{\vtheta_t}(a^h|s^h)}.
\end{align}
Here the importance sampling weight is introduced  to guarantee that $\vg(\tau_t;\vtheta_t) - \omega(\tau_t | \vtheta_{t-1}, \vtheta_t)\cdot \vg(\tau_t; \vtheta_{t-1})$ is an unbiased estimator of the gradient difference $\nabla J_\rho^H(\vtheta_{t}) - \nabla J_\rho^H(\vtheta_{t-1})$, i.e.,
\begin{align*}
	\E[\tau_{t}]{\vg(\tau_t;\vtheta_t) - \omega(\tau_t | \vtheta_{t-1}, \vtheta_t)\cdot \vg(\tau_t; \vtheta_{t-1})} =  \nabla J_\rho^H(\vtheta_{t}) - \nabla J_\rho^H(\vtheta_{t-1}).
\end{align*}
The momentum-based method combines  the unbiased SGD estimator~\cite{bottou2012stochastic} with the SARAH estimator~\cite{nguyen2017sarah} under a single-loop framework, which can avoid the high computational cost in variance reduction.


 Note, however, that the existence of the importance sampling weight requires a strong unverifiable assumption for the convergence analysis of the corresponding method. In order to handle this issue,  a Hessian-aided momentum-based estimator is proposed in \cite{fatkhullin2023stochastic, salehkaleybar2022momentum}. The main idea is to construct an unbiased estimate of $\nabla J_\rho^H({\vtheta}_t) - \nabla J_\rho^H({\vtheta}_{t-1})$ using the integral of policy Hessian. Let $\hbtheta_t$ be a linear combination of $\vtheta_{t}$ and $\vtheta_{t-1}$, 
\begin{align*}
	\hbtheta_t = q\vtheta_t + (1-q)\vtheta_{t-1}, \numberthis\label{eq:theta_m}
\end{align*}
where $q$ obeys the uniform distribution over $[0,1]$, denoted $\Unif(0,1)$. Let  $\hat{\tau}_t$ be a trajectory generated by the policy $\pi_{\hbtheta_t}$, and $\mH(\hat{\tau}_t; \hbtheta_t)$ be an unbiased estimate of policy Hessian $\nabla^2J_\rho^H (\hbtheta_t)$. 
A simple calculation yields that
\begin{align*}
\nabla J_\rho^H(\vtheta_{t}) - \nabla J_\rho^H(\vtheta_{t-1}) &= \int_0^1 \nabla^2 J_\rho^H\left(q\vtheta_t+ (1-q) \vtheta_{t-1}\right) (\vtheta_{t}-\vtheta_{t-1})dq\\
&=\E[q]{\nabla^2 J_\rho^H(\hbtheta_t)(\vtheta_{t}-\vtheta_{t-1})} \\
&=\E[q]{ \E[\hat{\tau}_t]{\mH(\hat{\tau}_t; \hbtheta_t)(\vtheta_{t}-\vtheta_{t-1})}\bigg| q}\\
&= \E{\mH(\hat{\tau}_t; \hbtheta_t)(\vtheta_{t}-\vtheta_{t-1})},
\end{align*}  
which implies that $\mH(\hat{\tau}_t; \hbtheta_t)(\vtheta_{t}-\vtheta_{t-1})$ is an unbiased estimator of $\nabla J_\rho^H({\vtheta}_t) - \nabla J_\rho^H({\vtheta}_{t-1})$. Based on this observation,
 the  Hessian-aided momentum-based estimator is given by
\begin{align*} \numberthis\label{eq:hess es}
\vu_t =  \beta_t \vg(\tau_{t}; \vtheta_t)  + (1-\beta_t)\lb \vu_{t-1} + \mH(\hat{\tau}_{t}; \hbtheta_t)(\vtheta_{t}-\vtheta_{t-1})\rb,
\end{align*}
where $\tau_{t}\sim p_{\vtheta_t}, \hat{\tau}_{t}\sim p_{\hbtheta_t}$, and the momentum coefficient $\beta_t$ is time-varying. 

\section{NPG-HM and Its Global Convergence}
\label{sec:alg and res}
\subsection{NPG-HM}
\label{sec: NPG-HM Algorithm}
As already mentioned,  NPG-HM  is a sample-based NPG with Hessian-aided momentum variance reduction. A detailed description of the method is presented in Algorithm \ref{alg:hessian_npg}. More concretely, the estimates of policy gradient and Hessian are  calculated using the expressions \eqref{eq:grad1} and \eqref{eq:hess1}, respectively. The unbiased estimator of policy gradient is calculated  using \eqref{eq:hess es}, see line~7 of Algorithm~\ref{alg:hessian_npg}. As is pointed out in \cite{fatkhullin2023stochastic}, the Hessian-vector product in line 7 can be efficiently computed as follows:
\begin{align*}
	\mH(\hat{\tau}_{t}; \hbtheta_t) \vx = \la \nabla \log p_{ \hbtheta_t}(\hat{\tau}_{t}), \vx\ra \vg(\hat{\tau}_{t};\hbtheta_t) + \nabla \la \vg(\hat{\tau}_{t};\hbtheta_t), \vx\ra,
\end{align*}
where $\vx\in\mathbb{R}^d$ is an arbitrary vector. In line~9, the update direction $\vw_t$ is generated by NPG-SGD (see Algorithm \ref{alg:sgd}), which applies the SGD method to solve the following sub-problem:
\begin{align}
\label{subprolbme1}
\min_{\vw\in\R^d} \left\{ \frac{1}{2} \E[(s,a)\sim \td_{\rho, \pi_{\vtheta_t}} ]{ \left(\vw^{\top} \nabla\log \pi_{\vtheta_t}(a | s) \right)^2 }- \vw^\tran\vu_t\right\}.
\end{align}
Since $\vu_t$ is an unbiased estimator of $\nabla J_\rho(\vtheta_t)$, the update direction $\vw_t$ can be regarded as an approximation of $\vw_t^\ast$ by noting \eqref{sub-prob of npg}. 


\begin{algorithm}[ht!]
\caption{Natural Policy Gradient with Hessian-aided Momentum (NPG-HM)}
\label{alg:hessian_npg}
\KwIn{Number of iterations $T$, horizon $H$, and initial parameter ${\vtheta}_{1}\in \mathbb{R}^{d}$}
\For{$t=1,2,\cdots, T-1$}{
\eIf{$t=1$}{
Sample a trajectory $\tau_1$ from $p_{\vtheta_1}$ and compute $\vu_1= \vg(\tau_{1};\vtheta_1)$\;
    
}{
    
Sample $q_t$ from $\Unif(0,1)$ and calculate $\hbtheta_t =q_t\vtheta_t + (1-q_t)\vtheta_{t-1}$\;
    
Sample two trajectories $\tau_{t}$ and $\hat{\tau}_{t}$ from $p_{\vtheta_t}$ and $p_{\hbtheta_t}$, respectively\;
    
Calculate
\begin{align*}
    \vu_t =  \beta_t \vg(\tau_{t}; \vtheta_t)  + (1-\beta_t)\lb \vu_{t-1} + \mH(\hat{\tau}_{t}; \hbtheta_t)(\vtheta_{t}-\vtheta_{t-1})\rb
\end{align*}
\hspace{-0.23cm} with $\beta_t =20/(t+20)$ \;
}
    
$\vw_t = \text{NPG-SGD}(\td_{\rho, \pi_{\vtheta_t}}, \pi_{\vtheta_t}, \vu_t)$\;

$\vtheta_{t+1}=\vtheta_t + \alpha_t \vw_t$ with $\alpha_t=\alpha_0\beta_t^{1/2}$\;
}
    
\KwOut{$\vtheta_{T}$}
\end{algorithm}

\begin{algorithm}[ht!]
\caption{NPG-SGD}
\label{alg:sgd}
\KwIn{Distribution $\td_{\rho, \pi_{\vtheta}}$, policy $\pi_{\vtheta}$, gradient estimate $\vu$, and initial parameter $\vw^{0}\in \mathbb{R}^{d}$}
    
\For{$k=0,\cdots, K-1$}{
Sample a state-action pair $(s_k,a_k)$ from $\td_{\rho, \pi_{\vtheta}}$\;
Calculate
\begin{align*}
    \vw^{k+1} = \vw^k - \eta \left( \lb {\vw^k}^\tran \nabla \log\pi_{\vtheta}(a_k|s_k)\rb\nabla \log\pi_{\vtheta}(a_k|s_k) - \vu \right)
\end{align*}
\hspace{-0.23cm} with $\eta = 1/(4M_g)$\;
    
} 

\KwOut{$\vw_{t}= \frac{1}{K+1}\sum_{k=0}^K \vw^k$}
\end{algorithm}

\subsection{Global convergence of NPG-HM}
\label{sec: Theoretical Results}
In this section, we present the global convergence of NPG-HM for the Fisher-non-degenerate policy class given in Definition~\ref{assumption:FIM}. Another two standard and useful assumptions, Assumptions~\ref{assumption:log-density} and \ref{assumption:approfim},  will also be first introduced. It is worth noting  that Definition~\ref{assumption:FIM}, Assumption~\ref{assumption:log-density}, and Assumption~\ref{assumption:approfim} can be satisfied by certain Gaussian policies. We refer interested readers  to~\cite{papini2018stochastic,xu2019sample, fatkhullin2023stochastic, liu2020improved} for more details. 


\begin{definition}
\label{assumption:FIM}
For an initial distribution $\rho$, a policy $\pi_{\vtheta}$ with $\vtheta\in\R^d$ is considered to be Fisher-non-degenerate if there exists a positive constant $\mu_F$ such that the Fisher Information Matrix (FIM) $\mF(\vtheta)$ induced by $\pi_{\vtheta}$ and $\rho$ \textup{(}see \eqref{eq: fim def}\textup{)} satisfies $\mF(\vtheta) \succcurlyeq \mu_F \mI_d$.
\end{definition}

\begin{assumption}
\label{assumption:log-density}
Let $\pi_{\vtheta}$ be the policy parameterized by $\vtheta\in\mathbb{R}^d$. There exist constants $M_g,M_h>0$ such that the gradient and Hessian of  the log-density of the policy satisfy
\begin{align*}
\| \nabla \log \pi_{\vtheta}(a|s) \|_2^2 \le M_g \text{ and } 
\| \nabla^2 \log \pi_{\vtheta}(a|s) \|_2 \le M_h
\end{align*}
for any $a\in\calA$ and $s\in \calS$.
\end{assumption}
\begin{remark}
Note that Assumption~\ref{assumption:log-density} is  generic and standard for  convergence analysis of policy gradient methods, see \textup{\cite{papini2018stochastic, xu2019sample, huang2020momentum, fatkhullin2023stochastic}}.  Moreover,  based on Definition~\ref{assumption:FIM} and Assumption~\ref{assumption:log-density}, we can easily conclude that
\begin{align*}
\mu_F \mI \preccurlyeq \mF(\vtheta) \preccurlyeq M_g \mI.
\end{align*}
In the sequel, we will let $\kappa = M_g/\mu_F$.     
\end{remark}


As the class of Fisher-non-degenerate policies may not encompass all stochastic policies, we will utilize the framework of compatible function approximation error, see \cite{agarwal2021theory, ding2021global, liu2020improved, fatkhullin2023stochastic}, to reflect the expressiveness of the policy class.

\begin{assumption}\label{assumption:approfim}
For any policy $\pi_{\vtheta}$ with some $\vtheta$, there exists a positive constant $\varepsilon_{\bias}$ such that
\begin{align*}
\E[(s,a)\sim \td_{\rho, \pi^\ast}]{\lb A^{\pi_{\vtheta}}(s,a)-(1-\gamma)(w^\ast_{\vtheta})^\tran\nabla\log\pi_{\vtheta}(a|s)\rb^2} \le \varepsilon_{\mathrm{bias}},
\numberthis\label{eq:cerror}
\end{align*}
where $\td_{\rho, \pi\ast}(s,a) = d_{\rho, \pi^\ast}(s)\cdot \pi^\ast(a|s)$ is the state-action distribution induced by an optimal policy $\pi^\ast$ that maximizes $J_\rho(\pi) = \E[s^0\sim\rho]{V^{\pi}(s^0)}$, and $\vw^\ast_{\vtheta} =\mF^{-1}(\vtheta)\nabla J_\rho(\vtheta)$  is the exact NPG update direction at $\vtheta$.
\end{assumption}

\begin{remark}
Assumption~\ref{assumption:approfim} states that the advantage function, denoted by $A^{\pi_{\vtheta}}(s,a)$, can be effectively approximated using the score function $\nabla \log \pi_{\vtheta}(a|s)$. Previous works have already established that $\varepsilon_{\bias}$ is zero for the tabular softmax parameterization policy \textup{\cite{agarwal2021theory}} or  the underlying MDP exhibiting a specific low-rank structure~\textup{\cite{jiang2017contextual, yang2019sample, jin2020provably}}. Furthermore, if the policy is parameterized using a neural network, $\varepsilon_{\mathrm{bias}}$ can be exceedingly small \textup{\cite{wang2019neural}}.  
\end{remark}


\begin{theorem}
\label{thm: main}
Suppose $H = -\frac{1}{\log \gamma} \log(T+\tau_0), \beta_t = \frac{\tau_0}{t+\tau_0}, \alpha_t = \alpha_0 \beta_t^{1/2}, \lambda_t = \lambda_0 \beta_t^{-1/2}, \lambda_0 =  \frac{\kappa \tau_0 \alpha_0}{4\mu_F}$ and $\alpha_0= \sqrt{\frac{\mu_F^2}{\kappa\tau_0(12L^2 + 6\nu_h^2)}}$, where $t\geq 1$ and $\tau_0\geq 20$. Assume NPG-SGD, see Algorithm~\ref{alg:sgd}, is run $K$ iterations to obtain an update direction with $K\geq 48\kappa^4(\sqrt{2d}+1)^2$. Let $\vtheta_1$ be the initialization. Then the output $\vtheta_T$ of Algorithm~\ref{alg:hessian_npg} obeys that
\begin{align*}
	\E{J_\rho^\ast - J_\rho(\vtheta_T)} & \lesssim  \frac{\E{J_\rho^\ast - J_\rho(\vtheta_1)}  }{T^2} + \frac{\nu_g^2}{T^2}  +\frac{\kappa^{3/2}}{1-\frac{\kappa^4 d}{K}  } \frac{1}{\sqrt{T}} + \frac{1}{T^{3/2}}+ \frac{\sqrt{\varepsilon_{\bias}}}{1-\gamma}.
\end{align*}
\end{theorem}

\begin{remark}
Theorem~\ref{thm: main} shows that NPG-HM enjoys an $\calO(\varepsilon^{-2})$ sample complexity to achieve the global last-iterate convergence. Note that the establishment of this result closely relies on the fact that it suffices to run the sub-problem solver \textup{(}i.e., Algorithm~\ref{alg:sgd}\textup{)} for a constant number of iterations \textup{(} i.e., the value of $K$ is independent of $\varepsilon$ to achieve $\E{J_\rho^\ast - J_\rho(\vtheta_T)}\lesssim \varepsilon$\textup{)}. This is indeed the key for us to 
be able to establish the $\calO(\varepsilon^{-2})$ sample complexity. 
In addition, a similar result can be established if we replace the gradient estimator in NPG-HM \textup{(}line 7 of Algorithm~\ref{alg:hessian_npg}\textup{)} by the one based on importance sampling, i.e., the one in \eqref{eq:momentum}. However, as already stated,  the analysis will then require an assumption on the importance sampling weight that is difficult to check. The details are omitted. 
\end{remark}

\section{Proof of Main Theorem}
\label{sec: proof of thm}
We first list some useful lemmas that will be used in the proof of Theorem~\ref{thm: main}. In particular, a new relaxed weak gradient dominance property and a novel ascent lemma are established, see Lemma~\ref{lemma gradient domination} and Lemma~\ref{asent lemma}, respectively. Moreover, the sample complexity of the subroutine (Algorithm~\ref{alg:sgd}) in  NPG-HM is carefully analyzed via an innovative error decomposition, see Lemma~\ref{lemma ascent 2}. Then a recursive sequence is developed in Lemma~\ref{lemma 3}.  The proofs of Lemmas \ref{lemma computational error}-\ref{lemma 3} are delayed to Section~\ref{sec: proofs of lmm}.

\begin{lemma}
\label{lmm:smooth}
Under Assumption \ref{assumption:log-density}, for all $\vtheta\in\R^d$, one has
\begin{itemize}
\item $J_\rho(\vtheta)$ is $L$-smooth, where $L=\frac{1}{(1-\gamma)^2} (M_g + M_h)$,
    \item $J_\rho^H(\vtheta)$ is $L$-smooth, where $L=\frac{1}{(1-\gamma)^2} (M_g + M_h)$, 
    \item $\max\left\{\twonorm{\nabla J_\rho^H(\vtheta)}, \twonorm{\nabla J_\rho(\vtheta)} \right\}\le \frac{\sqrt{M_g}}{(1-\gamma)^{3/2}}$.
\end{itemize}
\end{lemma}
\begin{remark}
The first result can be found in  \textup{\cite[Lemma 4.4]{yuan2022general}}, which improves the results  in \textup{\cite{xu2019sample, liu2020improved,papini2022smoothing}}. For the second one, the smoothness with a constant $L=\frac{M_h}{(1-\gamma)^2} + \frac{2M_g}{(1-\gamma)^3}$ is established   in \textup{\cite[Proposition 4.2]{xu2019sample}}. However, following a similar argument for the non-truncated objective \textup{(}i.e. proof for the first result in \textup{\cite[Lemma 4.4]{yuan2022general}}\textup{)}, it is evident that the constant can be improved to $L=\frac{1}{(1-\gamma)^2} (M_g + M_h)$.  The last one can be found in \textup{\cite[Lemma D.1]{yuan2022general}}, which improves the results in \textup{\cite{xu2019sample,liu2020improved,papini2022smoothing}}.
\end{remark}

\begin{lemma}
\label{lmm:variance}  
Let $\vg(\tau;\vtheta)$ and $\mH(\tau;\vtheta)$ be the unbiased estimates of policy gradient $\nabla J_\rho^H(\vtheta)$ and policy Hessian $\nabla^2 J_\rho^H(\vtheta)$, respectively.
Under Assumption~\ref{assumption:log-density} and the fact that $r(s,a, s')\in[-1,1]$, one has
\begin{align}
\label{eq g}
 \E[\tau]{\twonorm{\vg(\tau; \vtheta)-\nabla J^H_\rho (\vtheta)}^2}&\le \E[\tau]{\twonorm{\vg(\tau; \vtheta)}^2}\le \nu_g^2, \\
\label{eq H}
\E[\tau]{\opnorm{\mH(\tau; \vtheta)}^2} &\le \nu_h^2, 
\end{align}
where $\nu_g^2 = \frac{M_g}{(1-\gamma)^3}$ and $\nu_h^2 = \frac{2H^2M_g^2}{(1-\gamma)^3} + \frac{2M_h^2}{(1-\gamma)^4}$, and $\|\cdot\|$ is referred to as the spectral norm of a matrix. 
\end{lemma}
\begin{remark}
The variance of the gradient estimator  is provided in \textup{\cite[Lemma 4.2]{yuan2022general}}, which improves the results in \textup{\cite{shen2019hessian, pham2020hybrid, papini2022smoothing}}. In  \textup{\cite[Lemma 4.1]{shen2019hessian}}, an upper bound of $\E[\tau]{\opnorm{\mH(\tau; \vtheta)}^2}$ with $\nu_h^2 = \frac{2H^2M_g^2+2M_h^2}{(1-\gamma)^4}$ is established. However, based a better bound for $\E[\tau]{\twonorm{\vg(\tau; \vtheta)}^2}$ \textup{(}i.e., Equation~\eqref{eq g}\textup{)}, it is easy to see that $\nu_h^2$ can be improved the one presented in \textup{Lemma~\ref{lmm:variance}}.
\end{remark}


\begin{lemma}
\label{lmm:truncated}  
Under Assumption~\ref{assumption:log-density}, one has
\begin{align*}
\twonorm{\nabla J_\rho^H(\vtheta) - \nabla J_\rho(\vtheta)} &\le G_g\gamma^H, \\
\opnorm{\nabla^2 J_\rho^H(\vtheta) - \nabla^2 J_\rho(\vtheta)} & \le G_h\gamma^H,
\end{align*}
for any $\vtheta$, where $G_g = \frac{\sqrt{M_g}}{1-\gamma}\sqrt{\frac{1}{1-\gamma}+H}$, and
 $G_h = \frac{M_g+M_h}{1-\gamma}\lb \frac{1}{1-\gamma}+H \rb$.
\end{lemma}

\begin{remark} 
    The first result can be found in \textup{\cite[Lemma 4.5]{yuan2022general}}. The second result is presented in \textup{{\cite[Lemma 3]{masiha2022stochastic}}}, which also relies on a technical result in  \textup{\cite[Lemma B.6]{yuan2022general}}. Indeed, this technical result has been widely used in establishing  tighter bounds for the aforementioned quantities. 
\end{remark}

\begin{remark} 
    While the results in Lemma~\ref{lmm:smooth}, Equation~\eqref{eq g}, and Lemma~\ref{lmm:truncated} can be established under the following weaker conditions \textup{(}as is done for example in \textup{\cite{yuan2022general, papini2022smoothing}}\textup{)},
    \begin{align*}
       \E[a\sim \pi_{\vtheta}(\cdot|s)]{
       \twonorm{\nabla \log \pi_{\vtheta}(a|s)}^2} \le M_g \mbox{ and }\;
       \E[a\sim \pi_{\vtheta}(\cdot|s)]{\opnorm{ \nabla^2 \log \pi_{\vtheta}(a|s) }} \le M_h,
    \end{align*}
    it requires the stronger conditions in Assumption~\ref{assumption:log-density} to establish the bound for $\E[\tau]{\opnorm{\mH(\tau; \vtheta)}^2}$ \textup{(}i.e., Equation~\eqref{eq H}\textup{)}, to the best of our knowledge. 
\end{remark}

\begin{lemma}[\protect{\cite[Theorem 1]{bach2013non}}]
\label{Lemma bach2013non}
Suppose $\calH$ is a $d$-dimensional Euclidean space with $d\geq 1$. Let $(\vx_k,\vz_k)\in\calH \times\calH$ be independent and identically distributed observations. Assume that
\begin{itemize}
    \item $\E{\twonorm{\vx_k}^2}$ and $\E{\twonorm{\vz_k}^2}$ are finite, and $\E{\vx_k\vx_k^\tran} $ is invertible.
    \item The global minimum of $f(\vw) = 1/2\E{\la \vw,\vx_k \ra^2 - 2\la \vw, \vz_k\ra}$ is attained at a certain $\widehat{\vw}\in\calH$. Let $\xi_k = \vz_k - \la\widehat{\vw}, \vx_k \ra\vx_k$  be the residual, which obeys that $\E{\xi_k}=\bzero$. There exist $R>0, \sigma>0$ such that $\E{\xi_k\xi_k^\tran} \preceq \sigma^2 \E{\vx_k\vx_k^\tran}$ and $\E{\twonorm{\vx_k}^2 \vx_k\vx_k^\tran } \preceq R^2\E{\vx_k\vx_k^\tran}$.
    Consider the stochastic gradient decent recursion defined as
    \begin{align*}
        \vw^{k+1} = \vw^k - \eta (\la \vw^k, \vx_k\ra \vx_k - \vz_k),
    \end{align*}
    started from $\vw^0\in\calH$, where $\gamma>0$. Then for a constant step size $\eta = \frac{1}{4R^2}$, we have
    \begin{align*}
        \E{ f(\bar{\vw}_K) - f(\hw)} \leq \frac{2}{K} \left( \sigma \sqrt{d} + R \twonorm{\vw^0 - \hw}\right)^2,
    \end{align*}
    where $\bar{\vw}_K=(K+1)^{-1}\sum_{k=0}^{K} \vw^{k}$.
\end{itemize}
\end{lemma}

Lemma~\ref{Lemma bach2013non} enables us to analyse the convergence of the subroutine for computing the NPG update direction (Algorithm~\ref{alg:sgd}).

\begin{lemma}
\label{lemma computational error}
Given a pair of parameters $(\vu, \vtheta)$,  define the following loss function:
\begin{align*}
    \ell(\vw) = \frac{1}{2} \E[(s,a)]{  \left( \vw^\tran \nabla \log \pi_{\vtheta}(a|s) \right)^2 }  - \vw^\tran \vu,
\end{align*}
whose the minimizer is given by $\widehat{\vw}(\vu, \vtheta) = \mF^{-1}(\vtheta) \vu\in\R^d$.  Given a sampled pair $(s_k,a_k)\sim \td_{\rho, \pi_{\vtheta}}$ obtained as in  Algorithm~\ref{alg:sgd}, the unbiased gradient of $\ell(\vw)$ at $\vw^{k}$ is given by
\begin{align*}
\vg_{k}  =   \left( {\vw^{k}}^\tran \nabla \log \pi_{\vtheta}(a_k|s_k) \right) \nabla \log \pi_{\vtheta}(a_k|s_k) -  \vu.
\end{align*}
Consider the SGD procedure to solve this regression problem:
\begin{align*}
    \vw^{k+1} = \vw^{k} - \eta\vg_{k} \quad\text{for }k=0,\cdots, K-1,
\end{align*}
where  $\eta = \frac{1}{4M_g}$ and $\vw^{0} = \bzero$. Let $\vw_{o}(\vu, \vtheta)= (K+1)^{-1}\sum_{k=0}^{K} \vw^{k}$. Then one has
\begin{align}
    \E[(s_k,a_k)_{k=0,\cdots, K-1}]{\twonorm{ \vw_o(\vu,\vtheta) - \widehat{\vw}(\vu, \vtheta) }^2}  &\leq \frac{4M_g(\sqrt{2d}+1)^2}{K\mu_F^3} \twonorm{\vu}^2.
\end{align}
\end{lemma}

The next lemma provides a relaxed
weak gradient dominance property tailored for NPG under the Fisher-non-degenerate policy parameterization.

\begin{lemma}
\label{lemma gradient domination}
Under Definition~\ref{assumption:FIM}, Assumption~\ref{assumption:log-density}, and Assumption~\ref{assumption:approfim}, one has
\begin{align*}
\twonorm{\vw^\ast_{\vtheta}}^2 \geq \frac{1}{2M_g}  (J_\rho^\ast - J_\rho(\vtheta))^2 -  \frac{ \varepsilon_{\bias}}{M_g(1-\gamma)^2},
\end{align*}
where $\vw^\ast_{\vtheta} = \mF^{-1}(\vtheta)\nabla J_\rho(\vtheta)$.
\end{lemma}


\begin{lemma}
\label{asent lemma}
Suppose $\alpha_t\leq \frac{\mu_F}{2L}$. Under Definition~\ref{assumption:FIM} and Assumption \ref{assumption:log-density}, one has
\begin{align*}
J_\rho(\vtheta_{t+1})  &\geq J_\rho(\vtheta_t) +\frac{\alpha_t \mu_F}{2}   \twonorm{\vw_t^\ast}^2 + \frac{\alpha_t \mu_F}{4} \twonorm{\vw_t}^2 - \frac{2\alpha_tM_g}{\mu_F^2} G_g^2 \gamma^{2H} \\
 &\qquad -  \frac{2\alpha_tM_g}{\mu_F^2}\twonorm{ \nabla J_\rho^{H}(\vtheta_t) - \vu_t}^2 -  \alpha_tM_g  \twonorm{\vw_t - \hw_t}^2
\end{align*}
for any integer $t\geq 1$, where $\vw_t^\ast = \mF^{-1}(\vtheta_t)\nabla J_{\rho}(\vtheta_t)$, $\widehat{\vw}_t=\mF^{-1}(\vtheta_t)\vu_t$, and $\vw_t$ is the output of Algorithm~\ref{alg:sgd}.
\end{lemma}
\begin{remark}
Lemma~\ref{asent lemma} is derived especially for the NPG-type update \textup{(}i.e., line~10 of Algorithm~\ref{alg:hessian_npg}\textup{)} and is substantially different from that  for the PG-type update, as presented in \textup{\cite[Lemma 8]{fatkhullin2023stochastic}}. Additionally, compared with~\textup{{\cite[Proposition 4.5]{liu2020improved}}}, Lemma~\ref{asent lemma} enables us to establish the global last-iterate convergence of NPG-HM. Moreover, we can combine Lemma \ref{lemma gradient domination} and Lemma \ref{asent lemma} together to derive a recursive sequence on $J_\rho^\ast - \E{J_\rho(\vtheta_t)}$, thereby achieving an improved sample complexity.
\end{remark}

For the sake of simplicity, we will use the following shorthand notation in the remainder of this paper:
\begin{align*}
\Delta_t &= \E{J_\rho^\ast - J_\rho(\vtheta_t)},\\ 
R_t &= \E{\twonorm{\vtheta_{t+1} - \vtheta_t}^2} = \alpha_t^2 \E{\twonorm{\vw_t}^2},\\
V_t &= \E{\twonorm{ \nabla J_\rho^{H}(\vtheta_t) - \vu_t}^2},
\end{align*}
where $J_\rho^\ast=J_\rho(\pi^\ast)$ is the optimal objective function.

\begin{lemma}
\label{lemma ascent 2}
Let $c_1 = 1-\frac{24\kappa^4(\sqrt{2d}+1)^2}{K}$. Suppose $K\geq 48\kappa^4(\sqrt{2d}+1)^2$ and $\alpha_t\leq \frac{\mu_F}{2L}$. Under Definition~\ref{assumption:FIM} and Assumption \ref{assumption:log-density}, for every $t\geq 1$, one has
\begin{align*}
	\Delta_{t+1} \leq \Delta_t - \frac{c_1\alpha_t}{4\kappa} \Delta_t^2  + \frac{4\alpha_tM_g}{\mu_F^2}  V_t  - \frac{\mu_F}{4\alpha_t} R_t +  \frac{ \alpha_t \varepsilon_{\bias}}{2\kappa (1-\gamma)^2} + \frac{4\alpha_tM_g}{\mu_F^2} G_g^2 \gamma^{2H}.
\end{align*}
\end{lemma}


\begin{lemma}
\label{lemma: variance V}
Consider the sequence $\vu_t$  generated for every  integer $t\geq 1$ via  
\begin{align*}
\vu_t = (1-\beta_t) \left( \vu_{t-1} + \mH( \hat{\tau}_{t}; \hat{\vtheta}_t )(\vtheta_t - \vtheta_{t-1})\right)  + \beta_t \vg(\tau_{t}; \vtheta_t).
\end{align*}
Suppose $\beta_t\leq 1$. Under Assumption~\ref{assumption:log-density}, one has
\begin{align*}
    V_{t+1} \leq (1-\beta_{t+1})V_{t} + \left(12L^2 + 6\nu_h^2\right) R_{t} + 2\beta_{t+1}^2 \nu_g^2.
\end{align*}
\end{lemma}

\begin{lemma}
\label{lemma 3}
Suppose $\beta_t = \frac{\tau_0}{t+\tau_0}, \alpha_t = \alpha_0 \beta_t^{1/2}, \lambda_t = \lambda_0 \beta_t^{-1/2}, \lambda_0 =  \frac{\kappa \tau_0 \alpha_0}{4\mu_F}$ and $\alpha_0= \sqrt{\frac{\mu_F^2}{\kappa\tau_0(12L^2 + 6\nu_h^2)}}$, where $t\geq 1, \tau_0\geq 20$.  Define $\Lambda_t = \Delta_t + \lambda_{t-1} V_t$. Then $\Delta_t \leq \frac{2\sqrt{\varepsilon_{\bias}}}{1-\gamma}$ or 
\begin{align*}
	\Lambda_{t+1} \leq  \left(1 -\frac{2\beta_t}{\tau_0} \right) \Lambda_t + \left( \frac{16\kappa }{c_1 \alpha_0\tau_0^2}  + \frac{\tau_0 \kappa\alpha_0 \nu_g^2}{2\mu_F} \right)\beta^{3/2}_t + \frac{4\alpha_0 M_gG_g^2 \gamma^{2H}  }{\mu_F^2}  \beta_t^{1/2}.
\end{align*}
\end{lemma}

We are in position to present the proof of Theorem~\ref{thm: main}.

\paragraph{Proof of Theorem~\ref{thm: main}.}
Define $C_1 =  \frac{16\kappa }{c_1 \alpha_0\tau_0^2}  + \frac{\tau_0 \kappa\alpha_0 \nu_g^2}{2\mu_F} $ and $C_2 = \frac{4\alpha_0 M_gG_g^2 \gamma^{2H}  }{\mu_F^2}$. By Lemma \ref{lemma 3},  one has $\Delta_t \leq \frac{2\sqrt{\varepsilon_{\bias}}}{1-\gamma}$ or
\begin{align*}
	(t+\tau_0)^2 \Lambda_{t+1} &\leq (t+\tau_0)^2  \left(1 -\frac{2\beta_t}{\tau_0} \right) \Lambda_t + C_1 (t+\tau_0)^2  \beta^{3/2}_t +C_2 (t+\tau_0)^2  \beta_t^{1/2}\\
	&= (t+\tau_0)^2  \left(1 -\frac{2}{\tau_0} \cdot \frac{\tau_0}{t+\tau_0} \right) \Lambda_t + C_1 (t+\tau_0)^2  \beta^{3/2}_t +C_2 (t+\tau_0)^2  \beta_t^{1/2}\\
	&= (t+\tau_0)^2 \frac{t-2+\tau_0}{t+\tau_0} \Lambda_t + C_1 (t+\tau_0)^2  \beta^{3/2}_t +C_2 (t+\tau_0)^2  \beta_t^{1/2}\\
	&=(t+\tau_0) (t+\tau_0-2)\Lambda_t + C_1 (t+\tau_0)^2  \beta^{3/2}_t +C_2 (t+\tau_0)^2  \beta_t^{1/2}\\
	&\leq (t+\tau_0-1)^2\Lambda_t + C_1 (t+\tau_0)^2  \beta^{3/2}_t +C_2 (t+\tau_0)^2  \beta_t^{1/2}.
\end{align*}
Summing above inequality from $t=1$ to $T-1$ yields that
\begin{align*}
	(T+\tau_0 -1)^2 \Lambda_T \leq \tau_0^2 \Lambda_1 + \sum_{t=1}^{T-1}C_1 (t+\tau_0)^2  \beta^{3/2}_t +\sum_{t=1}^{T-1}C_2 (t+\tau_0)^2  \beta_t^{1/2}.
\end{align*}
Thus we have
\begin{align*}
	\Lambda_T &\leq \frac{\tau_0^2 \Lambda_1}{(T+\tau_0-1)^2} + \frac{\sum_{t=1}^{T-1}C_1 (t+\tau_0)^2  \beta^{3/2}_t}{(T+\tau_0 -1)^2} +\frac{\sum_{t=1}^{T-1}C_2 (t+\tau_0)^2  \beta_t^{1/2}}{(T+\tau_0 -1 )^2}\\
	&=\frac{\tau_0^2 \Lambda_1}{(T+\tau_0-1)^2} + \frac{C_1 \tau_0^{3/2}}{(T+\tau_0 -1 )^2}\sum_{t=1}^{T-1} (t+\tau_0)^{1/2}   +\frac{C_2\tau_0^{1/2}}{(T+\tau_0 -1 )^2}\sum_{t=1}^{T-1} (t+\tau_0 )^{3/2}\\
	&\leq \frac{\tau_0^2 \Lambda_1}{(T+\tau_0 -1)^2} + \frac{C_1 \tau_0^{3/2}}{(T+\tau_0-1)^2}\cdot \frac{1}{1+1/2}(T+\tau_0)^{3/2} +\frac{C_2\tau_0^{1/2}}{(T+\tau_0-1)^2} \cdot \frac{1}{1+3/2} (T+\tau_0)^{5/2}\\
	&\leq \frac{\tau_0^2 }{(T+\tau_0-1)^2} \left( \E{J_\rho^\ast - J_\rho(\vtheta_1) + \lambda_0 \nu_g^2} \right) + \frac{C_1 \tau_0^{3/2}}{(T+\tau_0-1)^2}\cdot \frac{1}{1+1/2}(T+\tau_0)^{3/2} \\
 &\qquad +\frac{C_2\tau_0^{1/2}}{(T+\tau_0-1)^2} \cdot \frac{1}{1+3/2} (T+\tau_0)^{5/2},
\end{align*}
where the second inequality is due to $\sum_{t=1}^{T-1}(t+\tau_0)^{\alpha} \leq \int_{t=0}^{T}(t+\tau_0)^{\alpha}dt \leq \frac{1}{1+\alpha}(T+\tau_0)^{1+\alpha} $ for any $\alpha >0$. Since $\alpha_0=\sqrt{\frac{\mu_F^2}{\kappa\tau_0(12L^2 + 6\nu_h^2)}} \leq \frac{\mu_F}{2L} $ and $c_1 = 1-\frac{24\kappa^4(\sqrt{2d}+1)^2}{K} $, we have
\begin{align*}
	C_1 &= \frac{16\kappa }{c_1 \alpha_0\tau_0^2}  + \frac{\tau_0 \kappa\alpha_0 \nu_g^2}{2\mu_F} \leq \frac{16\kappa }{c_1\tau_0^2}\cdot \sqrt{\frac{\kappa\tau_0(12L^2 + 6\nu_h^2)}{\mu_F^2}}   + \frac{\tau_0 \kappa \nu_g^2}{2\mu_F} \cdot \frac{\mu_F}{2L} \lesssim\frac{ \kappa^{3/2}}{ 1-\frac{\kappa^4 d}{K}  } +  \frac{\tau_0 \kappa \nu_g^2}{L}. 
\end{align*}
Furthermore, since $H = -\frac{1}{\log \gamma} \log(T+\tau_0)$, we have $\gamma^{2H} = (T+\tau_0)^{-2}$ and $C_2= \frac{ 4\alpha_0M_g G_g^2}{\mu_F^2} (T+\tau_0)^{-2}$. It follows that
\begin{align*}
	\Delta_T & \lesssim  \max \left( \Lambda_T, \frac{\sqrt{\varepsilon_{\bias}}}{1-\gamma}\right)\\
	&\leq \Lambda_T + \frac{\sqrt{\varepsilon_{\bias}}}{1-\gamma}\\
	&\lesssim \frac{\E{J_\rho^\ast - J_\rho(\vtheta_1)}  }{T^2} + \frac{\nu_g^2}{T^2}  +\frac{\kappa^{3/2}}{1-\frac{\kappa^4 d}{K}  } \frac{1}{\sqrt{T}} + \frac{1}{T^{3/2}}+ \frac{\sqrt{\varepsilon_{\bias}}}{1-\gamma}.
\end{align*}


\section{Numerical Experiments}\label{sec: numerical experiments}
In this section, numerical experiments are conducted to evaluate the performance of NPG-HM. We compare  NPG-HM with other state-of-the-art policy gradient methods, such as momentum-based natural policy gradient (MNPG)~\cite{chen2022decentralized}, stochastic recursive variance reduced natural policy gradient (NPG-SRVR)~\cite{liu2020improved}, proximal policy optimization (PPO)~\cite{schulman2017proximal}, and Hessian-aided recursive policy gradient (HARPG)~\cite{fatkhullin2023stochastic}. The numerical experiments cover several continuous control tasks from OpenAI Gym~\cite{brockman2016openai} using MuJoCo-based simulators~\cite{todorov2012mujoco}, including Reacher, Hopper, Walker2d, InvertedPendulum, InvertedDoublePendulum, and HalfCheetah.

\paragraph{Network architecture.} For these continuous  tasks, we utilize a stochastic Gaussian
policy $\pi_{\vtheta} (\cdot|s) \sim \calN\lb\mu_{\vtheta}(s), \Sigma_{\vtheta}(s)\rb$, where the mean $\mu_{\vtheta}(s)$ and the diagonal covariance matrix $\Sigma_{\vtheta}(s)$ are modeled by a one-hidden-layer neural network of size 64 and equipped with ReLU as the activation function. The outputs $\mu_{\vtheta}(s)$ and $\Sigma_{\vtheta}(s)$ will be obtained through additional Tanh and Softplus layers, respectively. As for the policy gradient estimation, we use the truncated gradient estimator, see~\eqref{eq:grad1}, with a baseline:
\begin{align*}
\vg(\tau;\vtheta)=\sum_{h=0}^{H-1}\sum _{i=h}^{H-1}\lb \gamma^{i} r^{i}-b(s^h)\rb\nabla \log \pi_{{\vtheta}}(a^{h}|s^{h}),
\end{align*}
where $b(s^h)$, only relying on the current state $s^h$ to ensure unbiasedness, is introduced to further reduce variance. In our experiments, $b(s)$ is approximated by a value neural network using one hidden layer with 32 neurons and ReLU as the activation function. 

\paragraph{Training details.}  
We test each task in our experiments with 5 random seeds, and report the undiscounted average return against the number of time steps. For NPG-SRVR, we choose the batch size $N$, epoch size $m$, and minibatch size $B$ to be $10, 2$, and $3$, respectively. Regarding PPO, we set the clipped parameter to be $\varepsilon=0.2$ and the number of epochs to be $K=10$. In MNPG, the momentum coefficient $\beta$ is set to be $0.5$. As for NPG-HM and HARPG, the  momentum coefficient $\beta_t$ is set to be $\frac{20}{t+20}$ and $\frac{2}{t+2}$, respectively. To improve the performance, we use Adam~\cite{kingma2014adam} instead of SGD (Algorithm~\ref{alg:sgd}) to solve for the update direction in NPG-HM, MNPG as well as NPG-SRVR. The number of iterations of Adam is set to $K=10$, and the learning rate of Adam is $1\times10^{-3}$. Furthermore, other hyperparameters of all training algorithms are fine-tuned for a fair comparison. More  on the setup of our  implementations are detailed  in Table~\ref{tab: para}.

\begin{table}[ht!]
\centering
\caption{Setup of environments and hyper-parameter details.}\label{tab: para}
{\small
\begin{tabular}{c |c |c |c |c |c |c } 
\hline
 Environments & Reacher & Hopper & Walker2d & Pendulum & DoublePendulum & HalfCheetah\\
\hline
\hline
Horizon  & 50& $1000$& 1000 & 1000& 1000& 1000\\
Number of timesteps & $1\times 10^6$ & $1\times 10^7$& $1\times 10^7$& $1\times 10^7$& $1\times 10^7$& $1\times 10^7$\\
Discount factor & 0.99& 0.99& 0.99& 0.99& 0.99&  0.99\\
Value function learning rate & $2.5\times10^{-3}$ & $2.5\times10^{-3}$ & $2.5\times10^{-3}$& $2.5\times10^{-3}$ & $2.5\times10^{-3}$ &$2.5\times10^{-3}$ \\
NPG-HM initial learning rate & $2\times10^{-3}$ & $2\times10^{-3}$& $2\times10^{-3}$& $2\times10^{-3}$& $2\times10^{-3}$& $1\times10^{-3}$ \\
MNPG learning rate & $2.5\times10^{-3}$ & $2.5\times10^{-3}$& $2.5\times10^{-3}$& $2.5\times10^{-3}$& $2.5\times10^{-3}$& $1.5\times10^{-3}$ \\
NPG-SRVR learning rate &$2\times10^{-3}$ & $2\times10^{-3}$& $2.5\times10^{-3}$& $2\times10^{-3}$& $1\times10^{-3}$& $2\times10^{-3}$ \\
PPO learning rate &$2\times10^{-3}$ & $2\times10^{-3}$& $2\times10^{-3}$& $2\times10^{-3}$& $2\times10^{-3}$& $2\times10^{-3}$ \\
HARPG initial learning rate & $2\times10^{-3}$ & $2\times10^{-3}$& $2\times10^{-3}$& $2\times10^{-3}$& $2\times10^{-3}$& $1\times10^{-3}$ \\
\hline
\end{tabular}}
\end{table} 

\begin{figure}[ht!]
\centering
    \begin{subfigure}[]{0.33\textwidth}
         \includegraphics[width=1\linewidth]{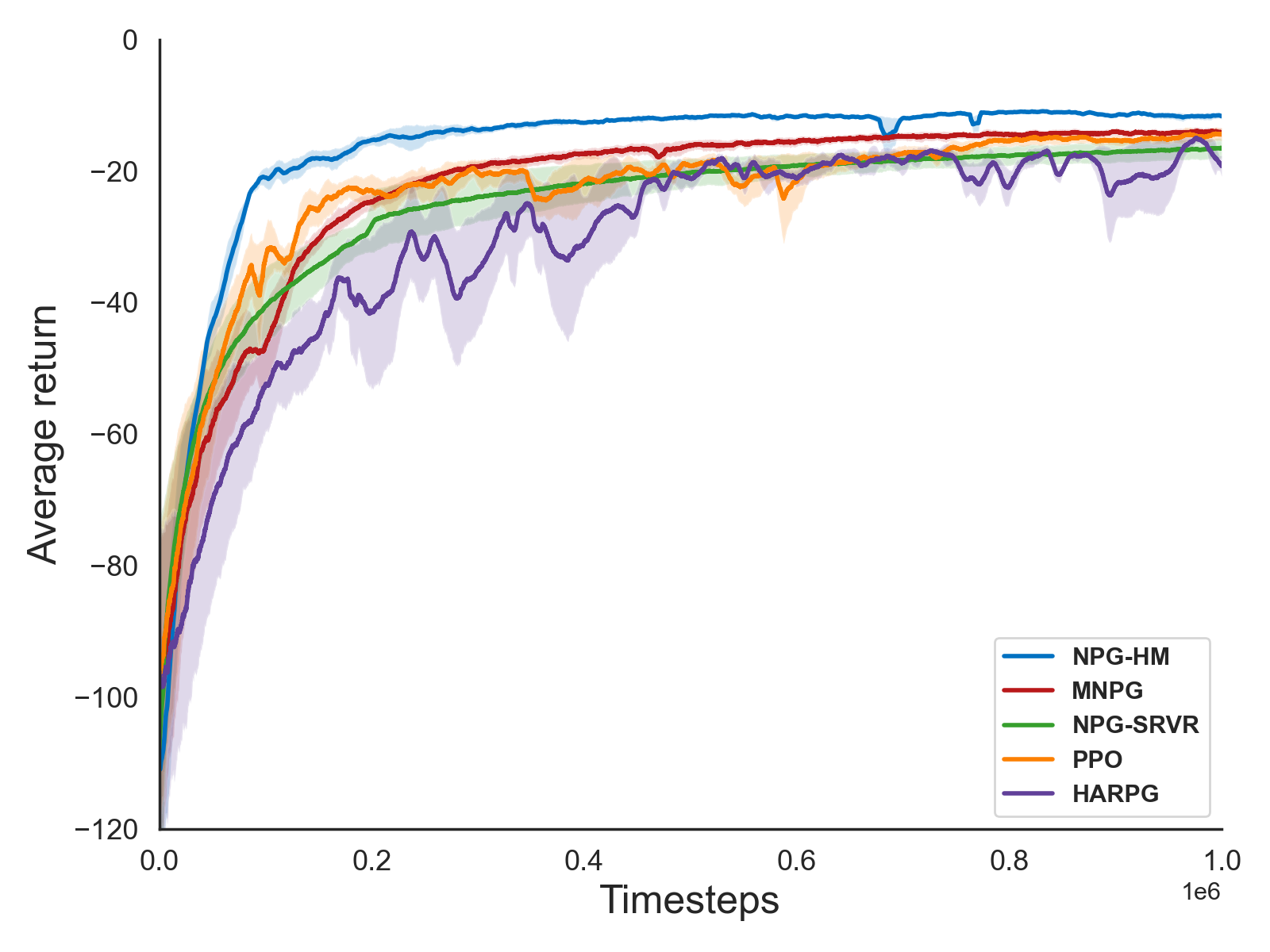}
         \caption[]{Reacher} 
         \label{fig: 1}
     \end{subfigure}
     \hfill \hspace{-20mm}
     \begin{subfigure}[]{0.33\textwidth}
         \includegraphics[width=1\linewidth]{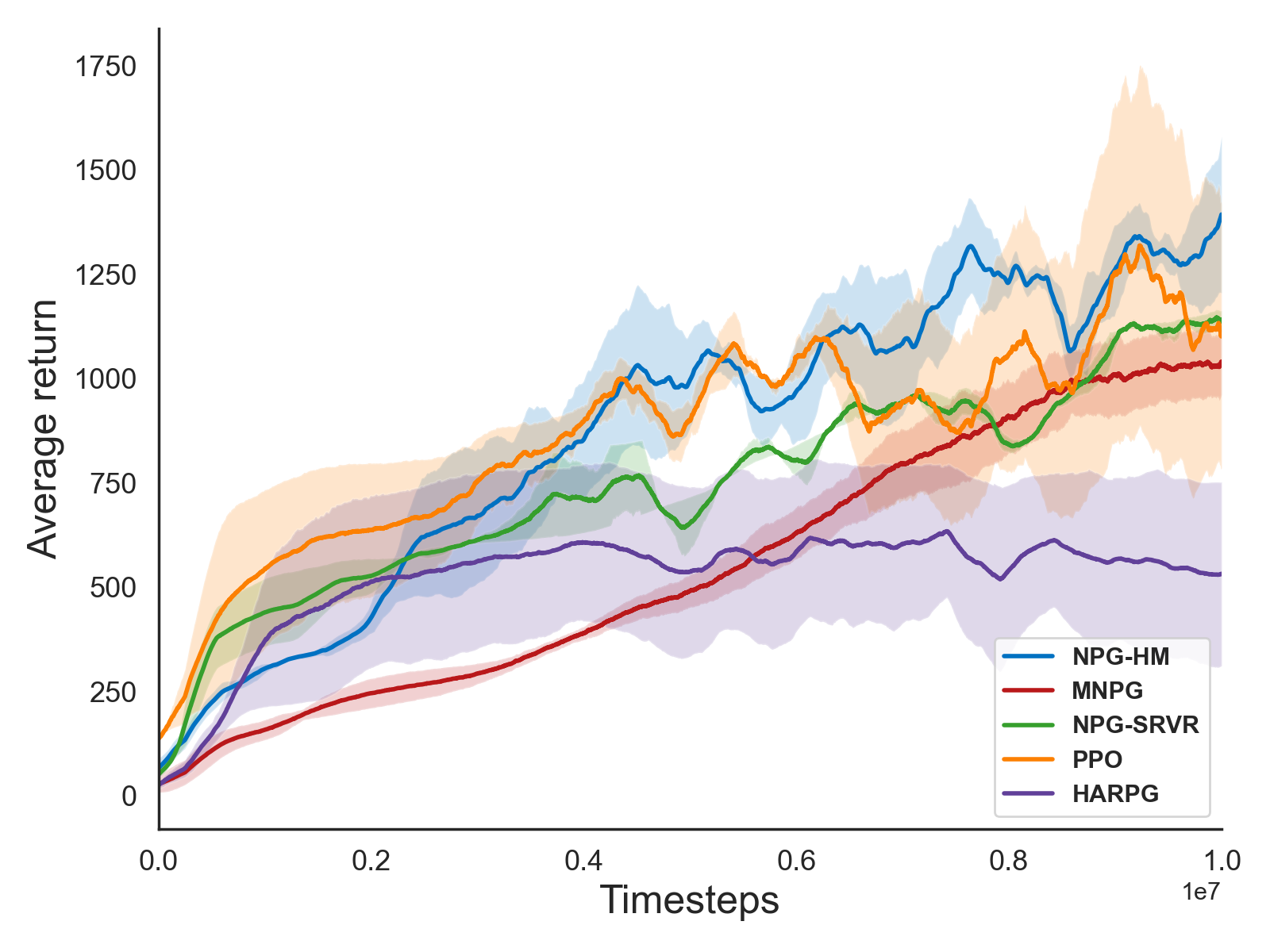}
         \caption[]{Hopper}  
         \label{fig: 2}
     \end{subfigure}
     \hfill  \hspace{-20mm}
    \begin{subfigure}[]{0.33\textwidth}
         \includegraphics[width=1\linewidth]{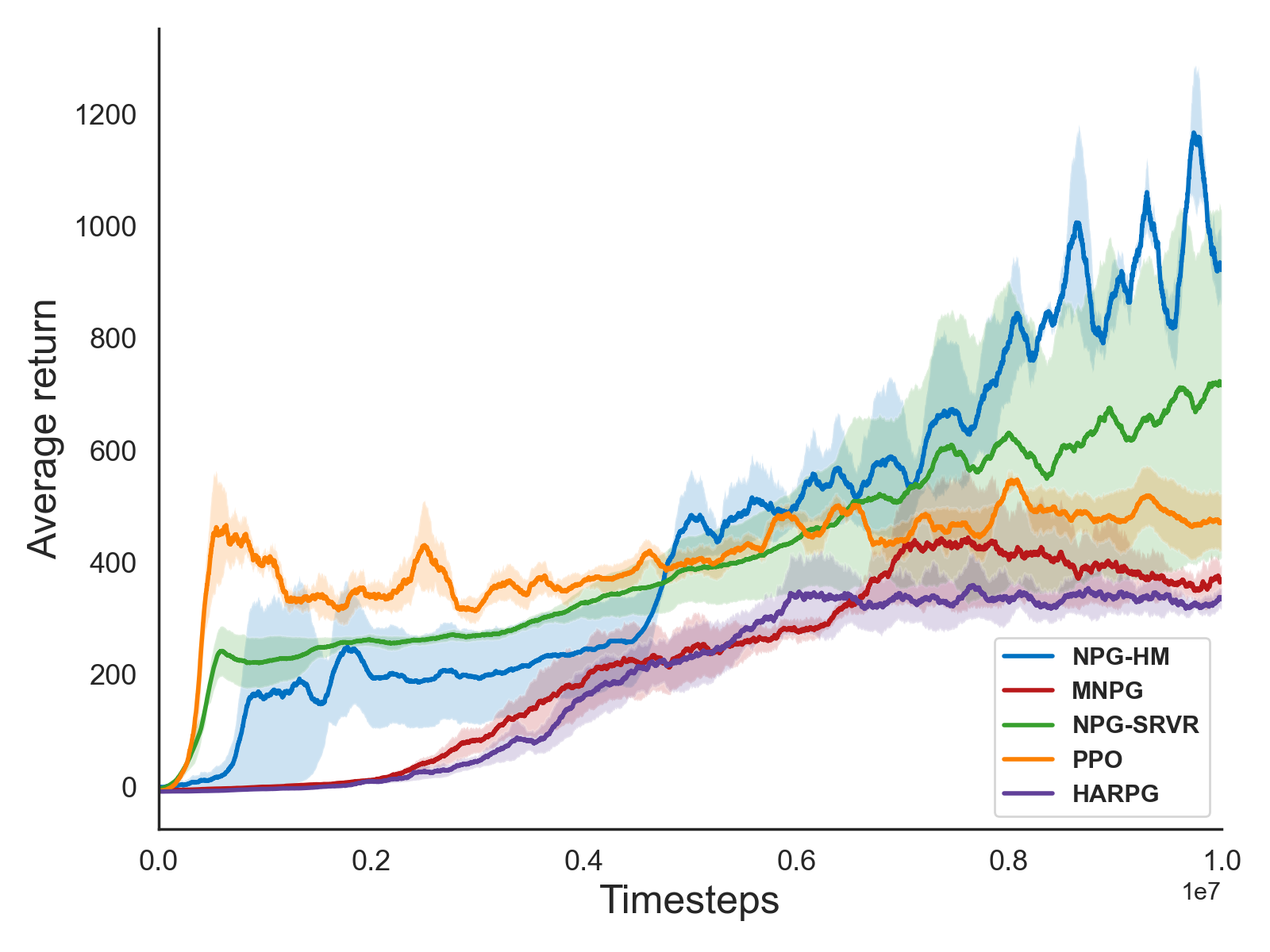}
         \caption[]{Walker2d}  
         \label{fig: 3}
     \end{subfigure}
     
     \vskip\baselineskip
     \begin{subfigure}[]{0.33\textwidth}
         \includegraphics[width=1\linewidth]{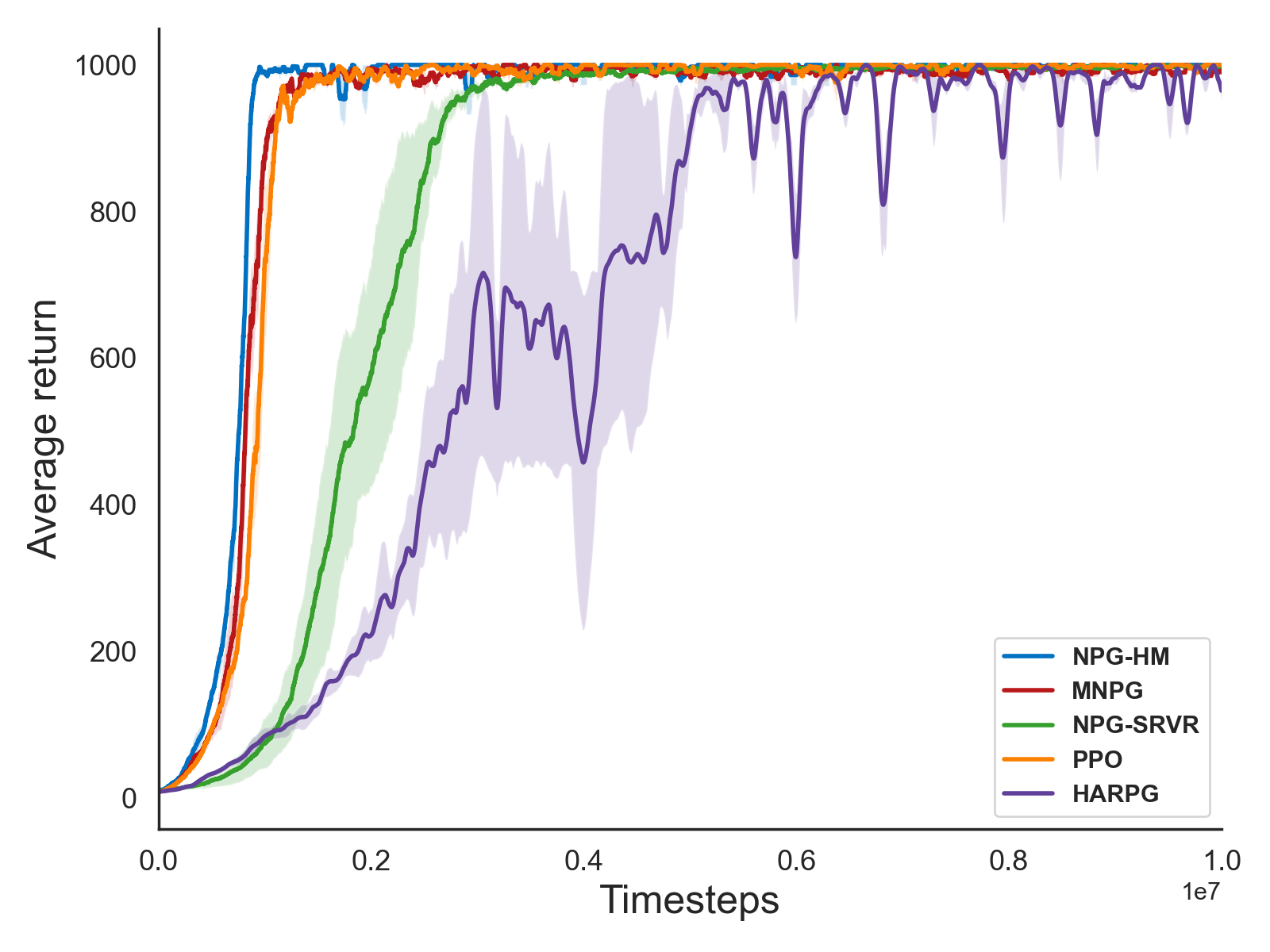}
         \caption[]{Pendulum}  
         \label{fig: 4}
     \end{subfigure}
     \hfill \hspace{-20mm}
    \begin{subfigure}[]{0.33\textwidth}
         \includegraphics[width=1\linewidth]{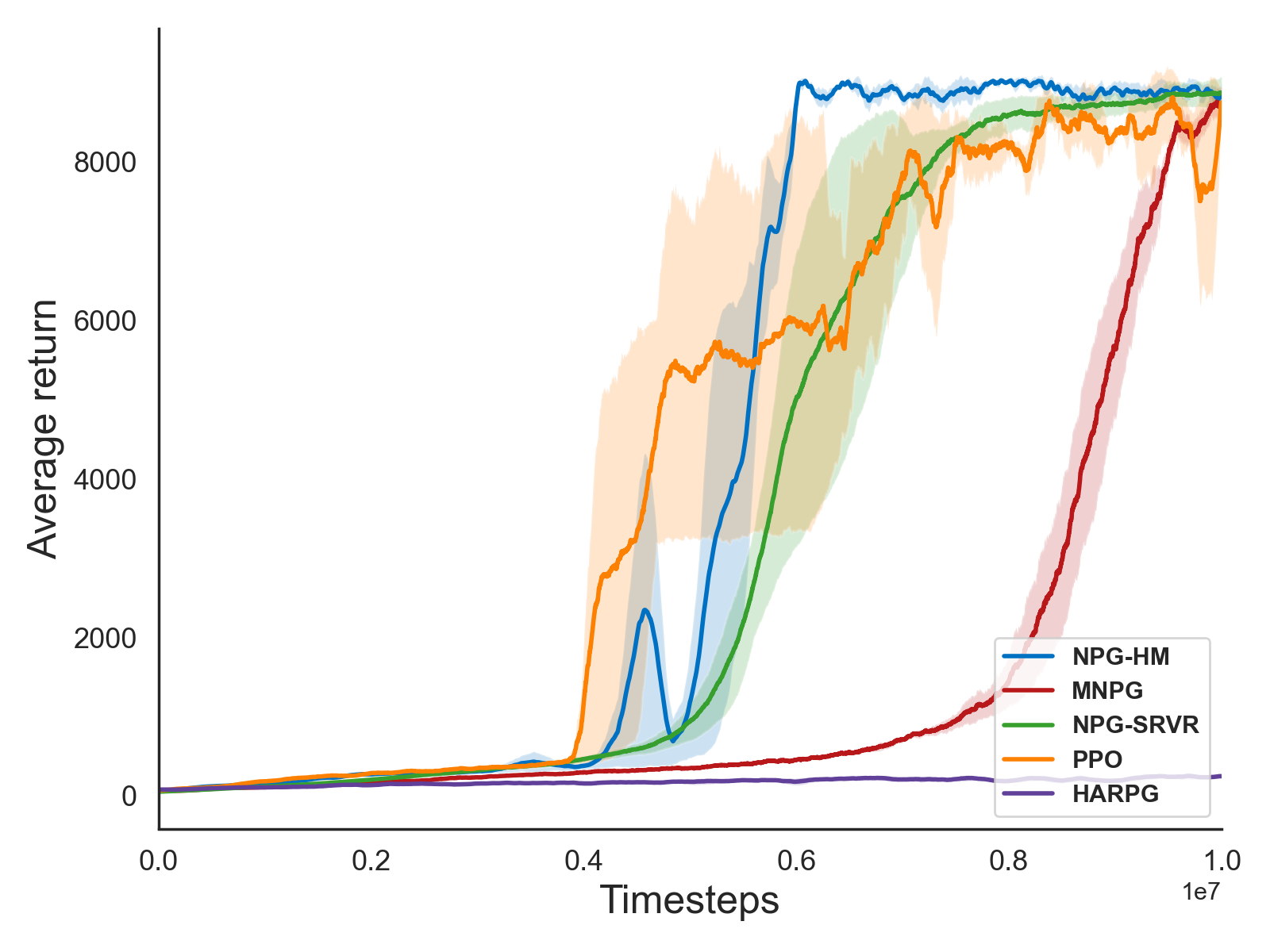}
         \caption[]{DoublePendulum} 
         \label{fig: 5}
     \end{subfigure}
     \hfill \hspace{-20mm}
     \begin{subfigure}[]{0.33\textwidth}
         \includegraphics[width=1\linewidth]{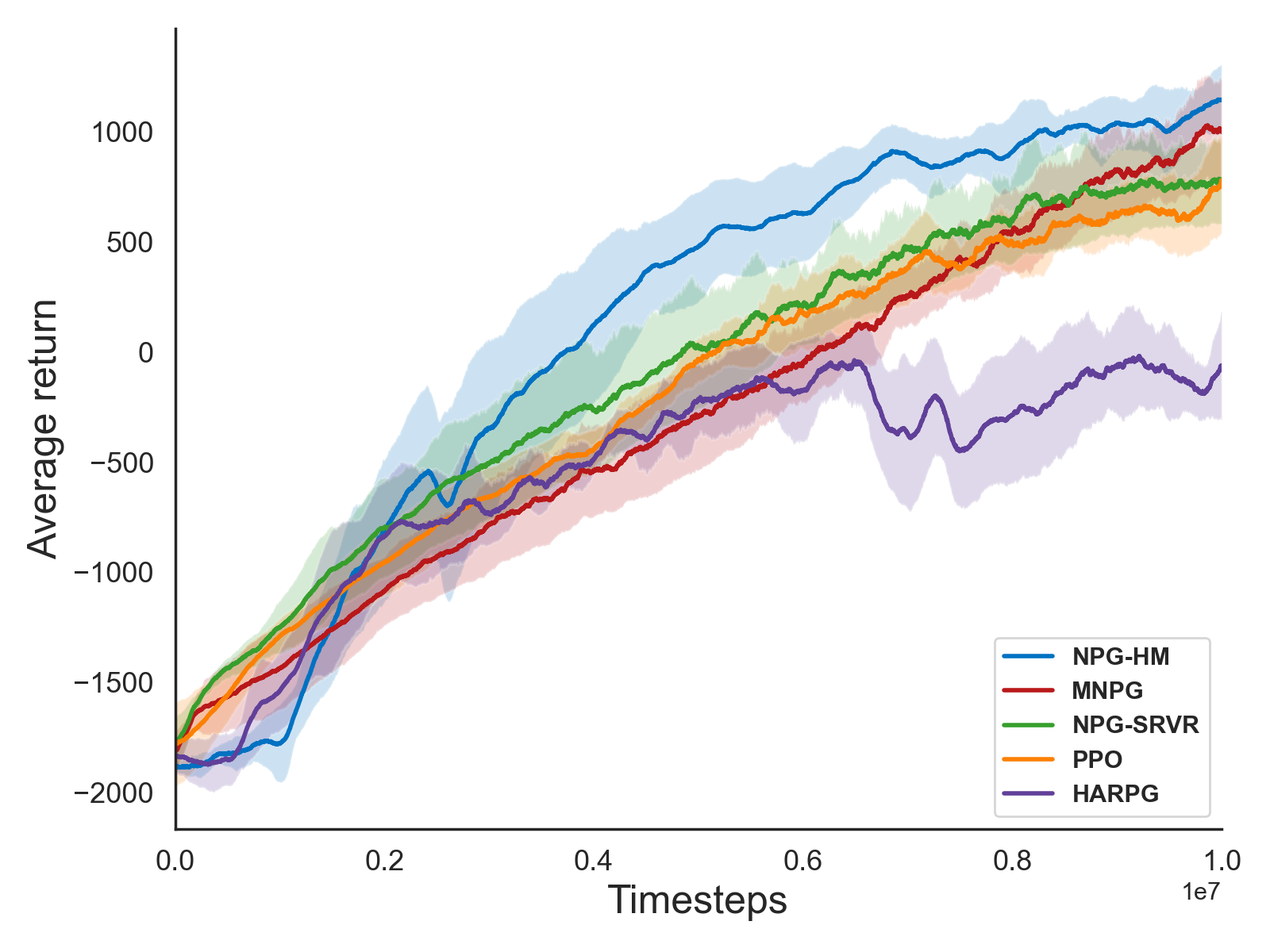}
         \caption[]{HalfCheetah} 
         \label{fig: 6}
     \end{subfigure}
    \caption{Empirical comparison of NPG-HM and other policy gradient methods on six environments.} 
    \label{fig: res}
\end{figure}

\paragraph{Performance comparison.}
The numerical results are displayed in Figure~\ref{fig: res}. Overall, NPG-HM exhibits the best performance among all the test methods. Regarding the comparisons with the other variance-reduced NPG methods, although NPG-SRVR performs similarly to  NPG-HM in the Walker2d environment, it is quite computationally expensive due to the double-loop scheme and two sampling batches at each iteration; and one potential reason why NPG-HM outperforms MNPG is that NPG-HM avoids the instability incurred by importance sampling weight. Moreover, NPG-HM performs significantly better than the  Hessian-aided momentum variance reduction PG method HARPG. As for the widely used  PPO method, the performance of  NPG-HM is still superior to it for the six tasks.

\section{Conclusions}\label{sec: conclusions}
In this paper, we have proposed a new algorithm termed NPG-HM for RL, which is a  sample-based NPG method with Hessian-aided moment variance reduction. The $\calO(\varepsilon^{-2})$ sample  complexity for NPG-HM to achieve the global last-iterate convergence has been established, which matches the best known result under the general Fisher-non-degenerate policy. Numerical results have been conducted to demonstrate the superiority of  NPG-HM compared to other state-of-the-art policy gradient methods.
There are several  directions for future research. Firstly, it is possible  to extend  NPG-HM  to the multi-agent setting and investigate the convergence properties. Furthermore, regularization-based NPG methods have been widely investigated in recent studies.  Therefore a possible future work is to study NPG-HM for the regularized RL problem. Lastly, it is also interesting to  design  more efficient subroutines to compute the update direction in  NPG-HM to improve the computational efficiency.

\section*{Acknowledgement}
The authors would like to thank Dr. Rui Yuan for pointing out that there are results with better constants for Lemmas~\ref{lmm:smooth}, \ref{lmm:variance}, and \ref{lmm:truncated}.

\section{Proofs of Technical Lemmas}\label{sec: proofs of lmm}
\subsection{Proof of Lemma \ref{lemma computational error}}
We will apply Lemma \ref{Lemma bach2013non} to prove this lemma.
Recall the definitions of $\vw_{o}(\vu, \vtheta)$ and $\widehat{\vw}(\vu, \vtheta)$ in the statement of the lemma.
Let $\vw_o = \vw_o(\vu, \vtheta), \hw = \hw(\vu, \vtheta)$, $\vx_k = \nabla \log \pi_{\vtheta}(a_k|s_k), \vz_k = \vu$ and $\xi_k = \vu- \la \hw, \vx_k\ra \vx_k$ for simplicity. A direct computation yields that
	\begin{align*}
		\twonorm{\vx_k}^2 &\leq M_g,\\
		\left( \hw^\tran\vx_k\right)^2 &\leq \twonorm{\vx_k}^2 \cdot \twonorm{\hw}^2 \leq M_g \twonorm{\mF^{-1}(\vtheta)\vu}^2 \leq \frac{M_g}{\mu_F^2} \twonorm{\vu}^2,\\
		\E[(s,a)]{\xi_k \xi_k^\tran} &\preccurlyeq \frac{M_g}{\mu_F^2} \twonorm{\vu}^2 \E[(s,a)]{\vx_k\vx_k^\tran} + \frac{\twonorm{\vu}^2}{\mu_F} \E[(s,a)]{\vx_k\vx_k^\tran}\\
		&\preccurlyeq \frac{2M_g}{\mu_F^2} \twonorm{\vu}^2 \E[(s,a)]{\vx_k\vx_k^\tran} :=\sigma^2 \E[(s,a)]{\vx_k\vx_k^\tran}, \\
		\E[(s,a)]{\twonorm{\vx_k}^2 \vx_k\vx_k^\tran} &\preccurlyeq M_g \E[(s,a)]{\vx_k\vx_k^\tran}:=R^2 \E[(s,a)]{\vx_k\vx_k^\tran},
	\end{align*}
	where $\sigma^2 = \frac{2M_g}{\mu_F^2} \twonorm{\vu}^2 $ and $R^2 = M_g$.
	Thus the application of Lemma \ref{Lemma bach2013non} yields that
	\begin{align*}
		\E{\twonorm{\vw_o - \hw }^2} &\leq \frac{2}{\mu_F}  \E{ f(\vw_o) - f(\hw) }\\
		&\leq \frac{2}{\mu_F} \cdot \frac{2}{K}  \left( \sigma \sqrt{d} + R \twonorm{\vw^0 - \hw}\right)^2\\
		&=\frac{2}{\mu_F} \cdot \frac{2}{K} \left( \sqrt{\frac{2M_g d}{\mu_F^2}} \twonorm{\vu}+ \sqrt{M_g}\twonorm{\hw}\right)^2\\
		&=\frac{2}{\mu_F} \cdot \frac{2}{K} \left( \sqrt{\frac{2M_g d}{\mu_F^2}} \twonorm{\vu}+ \sqrt{M_g}\twonorm{\mF^{-1}(\vtheta)\vu}\right)^2\\
		&\leq \frac{2}{\mu_F} \cdot \frac{2}{K} \left( \sqrt{\frac{2M_g d}{\mu_F^2}} \twonorm{\vu}+ \frac{\sqrt{M_g}}{\mu_F}\twonorm{\vu}\right)^2\\
		&=\frac{4M_g(\sqrt{2d}+1)^2}{K\mu_F^3} \twonorm{\vu}^2,
	\end{align*}
which completes the proof.

\subsection{Proof of Lemma \ref{lemma gradient domination}}
	By Assumption \ref{assumption:approfim}, we know that
	\begin{align*}
		\varepsilon_{\bias} &\geq \E[(s,a)\sim \td_{\rho, \pi^\ast}]{\left( A^{\pi_{\vtheta}}(s,a) - (1-\gamma) {\vw^\ast_{\vtheta}}^\tran \nabla \log \pi_{\vtheta}(a|s) \right)^2 } \\
		&\geq  \left(  \E[(s,a)\sim \td_{\rho, \pi^\ast}]{ A^{\pi_{\vtheta}}(s,a)} - (1-\gamma)  \E[(s,a)\sim \td_{\rho, \pi^\ast}]{ {\vw^\ast_{\vtheta}}^\tran \nabla \log \pi_{\vtheta}(a|s) } \right)^2\\
		&=(1-\gamma)^2 \left( J_\rho^\ast - J_\rho(\vtheta) - \E[(s,a)\sim \td_{\rho, \pi^\ast}]{ {\vw^\ast_{\vtheta}}^\tran \nabla \log \pi_{\vtheta}(a|s) } \right)^2,
	\end{align*}
	where the second line is due to Jensen's inequality and the last line follows from the performance difference lemma of \cite{kakade2002approximately}, i.e.,
 \begin{align*}
     J_\rho^\ast - J_\rho(\vtheta) = \frac{1}{1-\gamma} \E[(s,a)\sim \td_{\rho, \pi^\ast}]{A^{\pi_{\vtheta}}(s,a)}.
 \end{align*}
 After rearrangement one has
	\begin{align*}
		\frac{\sqrt{\varepsilon_{\bias}}}{1-\gamma} &\geq J_\rho^\ast - J_\rho(\vtheta) - \E[(s,a)\sim \td_{\rho, \pi^\ast}]{ {\vw^\ast_{\vtheta}}^\tran \nabla \log \pi_{\vtheta}(a|s) } \\
		&\geq J_\rho^\ast - J_\rho(\vtheta) - \sqrt{M_g} \twonorm{\vw^\ast_{\vtheta}},
	\end{align*}
	where the last line is due to Assumption \ref{assumption:log-density}. 
	It follows that
	\begin{align*}
		M_g \twonorm{\vw^\ast_{\vtheta}}^2+ \frac{ \varepsilon_{\bias}}{(1-\gamma)^2}&\geq \frac{1}{2} \left(  \sqrt{M_g} \twonorm{\vw^\ast_{\vtheta}} + \frac{\sqrt{\varepsilon_{\bias}}}{1-\gamma} \right)^2 \\
		&\geq \frac{1}{2} (J_\rho^\ast - J_\rho(\vtheta))^2,
	\end{align*}
	which completes the proof.

\subsection{Proof of Lemma \ref{asent lemma}}
Recall that $\vw_t$ is the output of Algorithm \ref{alg:sgd}, $\vw_{t}^\ast$ is a minimizer of \eqref{sub-prob of npg}, and $\hw_t$ is a minimizer of \eqref{subprolbme1}.
Since $J_\rho(\cdot)$ is $L$-smooth (Lemma \ref{lmm:smooth}), we have
\begin{align*}
	J_\rho(\vtheta_{t+1}) &\geq  J_\rho(\vtheta_t) + \la \nabla J_\rho(\vtheta_t), \vtheta_{t+1} - \vtheta_t \ra - \frac{L}{2}\twonorm{\vtheta_{t+1} - \vtheta_t}^2\\
	&\stackrel{(a)}{=}J_\rho(\vtheta_t) +\alpha_t \la \nabla J_\rho(\vtheta_t),  \vw_t\ra - \frac{L\alpha_t^2}{2}\twonorm{\vw_t}^2\\
	&\stackrel{(b)}{=}J_\rho(\vtheta_t) +\alpha_t \la  \mF(\vtheta_t) \vw_t^\ast,  \vw_t\ra - \frac{L\alpha_t^2}{2}\twonorm{\vw_t}^2\\
	&\stackrel{(c)}{=}J_\rho(\vtheta_t) +\frac{\alpha_t}{2} \la \mF(\vtheta_t) \vw_t^\ast, \vw_t^\ast \ra + \frac{\alpha_t}{2}\la \mF(\vtheta_t) \vw_t, \vw_t\ra \\
	&\quad - \frac{\alpha_t}{2}\la \mF(\vtheta_t) (\vw_t^\ast - \vw_t), \vw_t^\ast - \vw_t \ra - \frac{L\alpha_t^2}{2}\twonorm{\vw_t}^2\\
	&\geq J_\rho(\vtheta_t) +\frac{\alpha_t \mu_F}{2}   \twonorm{\vw_t^\ast}^2 + \frac{\alpha_t \mu_F}{2} \twonorm{\vw_t}^2 - \frac{\alpha_tM_g}{2} \twonorm{\vw_t^\ast - \vw_t }^2 - \frac{L\alpha_t^2}{2}\twonorm{\vw_t}^2\\
	&=J_\rho(\vtheta_t) +\frac{\alpha_t \mu_F}{2}   \twonorm{\vw_t^\ast}^2 + \left(\frac{\alpha_t \mu_F}{2}  -\frac{L\alpha_t^2}{2} \right)\twonorm{\vw_t}^2 - \frac{\alpha_tM_g}{2} \twonorm{\vw_t^\ast - \vw_t }^2 \\
	&\geq J_\rho(\vtheta_t) +\frac{\alpha_t \mu_F}{2}   \twonorm{\vw_t^\ast}^2 + \left(\frac{\alpha_t \mu_F}{2}  -\frac{L\alpha_t^2}{2} \right)\twonorm{\vw_t}^2 \\
	&\quad - \alpha_tM_g \twonorm{\vw_t^\ast - \hw_t }^2 -  \alpha_tM_g  \twonorm{\vw_t - \hw_t}^2\\
	&=J_\rho(\vtheta_t) +\frac{\alpha_t \mu_F}{2}   \twonorm{\vw_t^\ast}^2 + \left(\frac{\alpha_t \mu_F}{2}  -\frac{L\alpha_t^2}{2} \right)\twonorm{\vw_t}^2\\
	&\quad  - \alpha_tM_g \twonorm{\mF^{-1}(\vtheta_t) \left(\nabla J_\rho(\vtheta) - \vu_t \right)}^2 -  \alpha_tM_g  \twonorm{\vw_t - \hw_t}^2\\
	&\geq J_\rho(\vtheta_t) +\frac{\alpha_t \mu_F}{2}   \twonorm{\vw_t^\ast}^2 + \left(\frac{\alpha_t \mu_F}{2}  -\frac{L\alpha_t^2}{2} \right)\twonorm{\vw_t}^2 \\
	&\quad  - \frac{\alpha_tM_g}{\mu_F^2} \twonorm{\nabla J_\rho(\vtheta) - \vu_t}^2 -  \alpha_tM_g  \twonorm{\vw_t - \hw_t}^2\\
	&\geq J_\rho(\vtheta_t) +\frac{\alpha_t \mu_F}{2}   \twonorm{\vw_t^\ast}^2 + \left(\frac{\alpha_t \mu_F}{2}  -\frac{L\alpha_t^2}{2} \right)\twonorm{\vw_t}^2 \\
	&\quad - \frac{2\alpha_tM_g}{\mu_F^2} \twonorm{\nabla J_\rho(\vtheta) - \nabla J_\rho^{H}(\vtheta_t)}^2 -  \frac{2\alpha_tM_g}{\mu_F^2}\twonorm{ \nabla J_\rho^{H}(\vtheta_t) - \vu_t}^2 -  \alpha_tM_g  \twonorm{\vw_t - \hw_t}^2\\
	&\stackrel{(d)}{\geq} J_\rho(\vtheta_t) +\frac{\alpha_t \mu_F}{2}   \twonorm{\vw_t^\ast}^2 + \left(\frac{\alpha_t \mu_F}{2}  -\frac{L\alpha_t^2}{2} \right)\twonorm{\vw_t}^2 \\
	&\quad - \frac{2\alpha_tM_g}{\mu_F^2} G_g^2 \gamma^{2H} -  \frac{2\alpha_tM_g}{\mu_F^2}\twonorm{ \nabla J_\rho^{H}(\vtheta_t) - \vu_t}^2 -  \alpha_tM_g  \twonorm{\vw_t - \hw_t}^2\\
	&\geq J_\rho(\vtheta_t) +\frac{\alpha_t \mu_F}{2}   \twonorm{\vw_t^\ast}^2 + \frac{\alpha_t \mu_F}{4} \twonorm{\vw_t}^2\\
	&\quad  - \frac{2\alpha_tM_g}{\mu_F^2} G_g^2 \gamma^{2H} -  \frac{2\alpha_tM_g}{\mu_F^2}\twonorm{ \nabla J_\rho^{H}(\vtheta_t) - \vu_t}^2 -  \alpha_tM_g  \twonorm{\vw_t - \hw_t}^2,
\end{align*}
where step (a) is due to the update rule $\vtheta_{t+1} = \vtheta_t + \alpha_t \vw_t$, step (b) is due to the definition $\vw_t^\ast = \mF^{-1}(\vtheta_t)\nabla J_\rho(\vtheta_t)$, step (c) follows from $\va^\tran\mM \vb = \frac{1}{2}\left(\va^\tran\mM\va + \vb^\tran\mM\vb - (\va-\vb)^\tran\mM(\va-\vb)\right)$ for any $\va,\vb$ and any symmetric matrix $\mM$, step (d) holds by Lemma \ref{lmm:truncated},  and the last line is due to $\alpha_t\leq \frac{\mu_F}{2L}$.

\subsection{Proof of Lemma \ref{lemma ascent 2}}
	By Lemma \ref{asent lemma}, one has
	\begin{align*}
		\E{J_\rho(\vtheta_{t+1}) } &\geq \E{J_\rho(\vtheta_t)} +\frac{\alpha_t \mu_F}{2}   \E{\twonorm{\vw_t^\ast}^2} + \frac{\alpha_t \mu_F}{4}\E{ \twonorm{\vw_t}^2} - \frac{2\alpha_tM_g}{\mu_F^2} G_g^2 \gamma^{2H} \\
		&\quad -  \frac{2\alpha_tM_g}{\mu_F^2} \E{\twonorm{ \nabla J_\rho^{H}(\vtheta_t) - \vu_t}^2 } -  \alpha_tM_g \E{ \twonorm{\vw_t - \hw_t}^2}\\
		&= \E{ J_\rho(\vtheta_t) } +\frac{\alpha_t \mu_F}{2}  \E{ \twonorm{\vw_t^\ast}^2 }+ \frac{ \mu_F}{4\alpha_t} R_t - \frac{2\alpha_tM_g}{\mu_F^2} G_g^2 \gamma^{2H} \\
  &\qquad -  \frac{2\alpha_tM_g}{\mu_F^2}V_t -  \alpha_tM_g  \E{\twonorm{\vw_t - \hw_t}^2}. \numberthis\label{eq a}
	\end{align*}
	Notice that $\vw_t$ is the output of Algorithm \ref{alg:sgd} and $\hw_t$ is the minimizer of \eqref{subprolbme1} given $(\vu_t, \vtheta_t)$. Then conditioned  on $\vu_t$ and $\vtheta_t$, applying Lemma \ref{lemma computational error} with $(\vu = \vu_t, \vtheta = \vtheta_t)$ yields that
	\begin{align*}
		\E[(s_k,a_k)_{k=0,\cdots, K-1}]{\twonorm{\vw_t - \hw_t}^2}&=\E[(s_k,a_k)_{k=0,\cdots, K-1}]{\twonorm{ \vw_o(\vu_t,\vtheta_t) - \widehat{\vw}(\vu_t, \vtheta_t) }^2}  \\
		&\leq \frac{4M_g(\sqrt{2d}+1)^2}{K\mu_F^3} \twonorm{\vu_t}^2.
	\end{align*}
Taking expectation on both sides yields that
	\begin{align*}
		&\E{\twonorm{ \vw_{t} - \widehat{\vw}_t }^2}  \\
  &\leq \frac{4M_g(\sqrt{2d}+1)^2}{K\mu_F^3} \E{\twonorm{\vu_t}^2}\\
		&\leq \frac{12M_g(\sqrt{2d}+1)^2}{K\mu_F^3}  \left(\E{\twonorm{ \vu_t - \nabla J_\rho^{H}(\vtheta_t)}^2} + \E{\twonorm{\nabla J_\rho^{H}(\vtheta_t)- \nabla J_\rho(\vtheta_t) }^2} + \E{\twonorm{\nabla J_\rho(\vtheta_t)  }^2} \right) \\
		&= \frac{12M_g(\sqrt{2d}+1)^2}{K\mu_F^3}\left(  \E{\twonorm{ \vu_t - \nabla J_\rho^{H}(\vtheta_t)}^2}+ \E{\twonorm{\nabla J_\rho^{H}(\vtheta_t)- \nabla J_\rho(\vtheta_t) }^2}  + \E{\twonorm{\mF(\vtheta_t)\vw_t^\ast }^2}  \right) \\
		&\leq \frac{12M_g(\sqrt{2d}+1)^2}{K\mu_F^3} \left(  \E{\twonorm{ \vu_t - \nabla J_\rho^{H}(\vtheta_t)}^2}+ G_g^2 \gamma^{2H} + M_g^2\E{\twonorm{\vw_t^\ast }^2}   \right)\\
		&=\frac{12M_g(\sqrt{2d}+1)^2}{K\mu_F^3}  \left(  V_t+ M_g^2\E{\twonorm{\vw_t^\ast }^2}  + G_g^2 \gamma^{2H}  \right). \numberthis\label{eq b}
	\end{align*}
	After plugging \eqref{eq b} into \eqref{eq a}, one has
	\begin{align*}
		\E{J_\rho(\vtheta_{t+1}) } &\geq\E{ J_\rho(\vtheta_t) } +\frac{\alpha_t \mu_F}{2}  \E{ \twonorm{\vw_t^\ast}^2 }+ \frac{ \mu_F}{4\alpha_t} R_t - \frac{2\alpha_tM_g}{\mu_F^2} G_g^2 \gamma^{2H} -  \frac{2\alpha_tM_g}{\mu_F^2}V_t \\
		&\quad -  \alpha_tM_g  \left( \frac{12M_g(\sqrt{2d}+1)^2}{K\mu_F^3}    V_t +\frac{12M_g^3(\sqrt{2d}+1)^2}{K\mu_F^3} \E{\twonorm{\vw_t^\ast }^2}  + \frac{12M_g(\sqrt{2d}+1)^2}{K\mu_F^3} G_g^2 \gamma^{2H}   \right)\\
		&=\E{J_\rho(\vtheta_t) } + \left( \frac{\alpha_t \mu_F}{2} -\frac{12\alpha_tM_g^4(\sqrt{2d}+1)^2}{K\mu_F^3}  \right)  \E{ \twonorm{\vw_t^\ast}^2 } +  \frac{ \mu_F}{4\alpha_t} R_t \\
		&\quad - \left(\frac{2\alpha_tM_g}{\mu_F^2} + \frac{12\alpha_tM_g^2(\sqrt{2d}+1)^2}{K\mu_F^3} \right)G_g^2 \gamma^{2H}- \left( \frac{2\alpha_tM_g}{\mu_F^2}  + \frac{12\alpha_tM_g^2(\sqrt{2d}+1)^2}{K\mu_F^3} \right)V_t\\
		&=\E{J_\rho(\vtheta_t) } + \frac{\alpha_t \mu_F}{2}  \left( 1-\frac{24\kappa^4(\sqrt{2d}+1)^2}{K}  \right)  \E{ \twonorm{\vw_t^\ast}^2 } +  \frac{ \mu_F}{4\alpha_t}R_t \\
		&\quad - \frac{2\alpha_tM_g}{\mu_F^2}\left( 1+ \frac{ 6\kappa (\sqrt{2d}+1)^2}{K}\right)G_g^2 \gamma^{2H} - \frac{2\alpha_tM_g}{\mu_F^2}   \left( 1+ \frac{6\kappa(\sqrt{2d}+1)^2}{K} \right)V_t\\
		&\geq \E{J_\rho(\vtheta_t) } +  \frac{ c_1 \alpha_t \mu_F}{2} \E{ \twonorm{\vw_t^\ast}^2 } +  \frac{\mu_F}{4\alpha_t } R_t- \frac{4\alpha_tM_g}{\mu_F^2} G_g^2 \gamma^{2H}  -\frac{4\alpha_tM_g}{\mu_F^2} V_t, 
	\end{align*}
	where  the last line is due to $K\geq 48\kappa^4(\sqrt{2d}+1)^2$, i.e.
	\begin{align*}
		c_1 := 1-\frac{24\kappa^4(\sqrt{2d}+1)^2}{K}  \geq \frac{1}{2} \text{ and }c_2 :=1 + \frac{6\kappa(\sqrt{2d}+1)^2}{K} \leq 2.
	\end{align*}
	Furthermore, the application of Lemma~\ref{lemma gradient domination} yields that
	\begin{align*}
		\E{J_\rho(\vtheta_{t+1}) } &\geq \E{ J_\rho(\vtheta_t) }  + \frac{c_1\alpha_t \mu_F}{2}   \left( \frac{1}{2M_g}  \E{(J_\rho^\ast - J_\rho(\vtheta_t))^2} -  \frac{ \varepsilon_{\bias}}{M_g(1-\gamma)^2} \right)\\
  &\qquad +  \frac{\mu_F}{4\alpha_t } R_t - \frac{4\alpha_tM_g}{\mu_F^2} G_g^2 \gamma^{2H}  -\frac{4\alpha_tM_g}{\mu_F^2}  V_t\\
		&=\E{ J_\rho(\vtheta_t) }  + \frac{c_1\alpha_t}{4\kappa}   \E{(J_\rho^\ast - J_\rho(\vtheta_t))^2} -  \frac{c_1 \alpha_t \varepsilon_{\bias}}{2\kappa (1-\gamma)^2} 
		+  \frac{\mu_F}{4\alpha_t } R_t - \frac{4\alpha_tM_g}{\mu_F^2} G_g^2 \gamma^{2H}  -\frac{4\alpha_tM_g}{\mu_F^2}  V_t\\
		&\geq  \E{ J_\rho(\vtheta_t) }  + \frac{c_1\alpha_t}{4\kappa}   \left( \E{J_\rho^\ast - J_\rho(\vtheta_t)} \right)^2 -  \frac{\alpha_t \varepsilon_{\bias}}{2\kappa (1-\gamma)^2}+  \frac{\mu_F}{4\alpha_t } R_t - \frac{4\alpha_tM_g}{\mu_F^2} G_g^2 \gamma^{2H}  -\frac{4\alpha_tM_g}{\mu_F^2}  V_t,
	\end{align*}
	where the last line follows from the Jensen's inequality and the fact $c_1\leq 1$. Recall that $\Delta_t = \E{J_\rho^\ast - J_\rho(\vtheta_t)}$.  A simple computation yields that
	\begin{align*}
		\Delta_{t+1} \leq  \Delta_t - \frac{c_1\alpha_t}{4\kappa} \Delta_t^2  + \frac{4\alpha_tM_g}{\mu_F^2}  V_t  - \frac{\mu_F}{4\alpha_t} R_t +  \frac{ \alpha_t \varepsilon_{\bias}}{2\kappa (1-\gamma)^2} + \frac{4\alpha_tM_g}{\mu_F^2} G_g^2 \gamma^{2H},
	\end{align*}
	which completes the proof.

\subsection{Proof of Lemma \ref{lemma: variance V}}
Define 
\begin{align*}
	\calG_t &=   \vg(\tau_{t}; \vtheta_t) - \nabla J_\rho^H(\vtheta_t),\\
	\calH_t &= \nabla J_\rho^H(\vtheta_{t-1}) -  \nabla J_\rho^H(\vtheta_{t}) + \mH( \hat{\tau}_{t}; \hat{\vtheta}_t )(\vtheta_t - \vtheta_{t-1}).
\end{align*}
We know that $\E{\calG_t} = 0$ and $\E{\twonorm{\calG_t}^2}\leq \nu_g^2$. Moreover, one can verify that $\E{\calH_t}=\bzero$ as follows:
\begin{align*}
	\E{\calH_t} &=\E{ \nabla J_\rho^H(\vtheta_{t-1}) -  \nabla J_\rho^H(\vtheta_{t}) } + \E{\mH( \hat{\tau}_{t}; \hat{\vtheta}_t )(\vtheta_t - \vtheta_{t-1})}\\
	&=\E{ \nabla J_\rho^H(\vtheta_{t-1}) -  \nabla J_\rho^H(\vtheta_{t}) } + \E{\nabla^2 J_\rho^{H}(\hat{\vtheta}_t)(\vtheta_t - \vtheta_{t-1})}\\
	&=\E{ \nabla J_\rho^H(\vtheta_{t-1}) -  \nabla J_\rho^H(\vtheta_{t}) } + \E{ \int_{0}^{1}\nabla^2 J_\rho^{H}(\xi \vtheta_t + (1-\xi)\vtheta_{t-1})(\vtheta_t - \vtheta_{t-1}) d\xi}\\
	&=0.
\end{align*}
Furthermore, the variance of $\calH_t$ can be bounded as follows:
\begin{align*}
	\E{\twonorm{\calH_t}^2} &= \E{\twonorm{\nabla J_\rho^H(\vtheta_{t-1}) -  \nabla J_\rho^H(\vtheta_{t}) +  \mH( \hat{\tau}_{t}; \hat{\vtheta}_t )(\vtheta_t - \vtheta_{t-1})}^2}\\
	&\leq  3  \E{\twonorm{\nabla J_\rho^H(\vtheta_{t-1}) -  \nabla J_\rho^H(\vtheta_{t}) }^2} + 3\E{\twonorm{ \left( \mH( \hat{\tau}_{t}; \hat{\vtheta}_t ) - \nabla^2 J_\rho^{H}(\hat{\vtheta}_t) \right) (\vtheta_t - \vtheta_{t-1})}^2}  \\
	&\quad + 3\E{\twonorm{ \nabla^2 J_\rho^{H}(\hat{\vtheta}_t) (\vtheta_t - \vtheta_{t-1}) }^2}\\
	&\leq  6L^2 \E{\twonorm{\vtheta_t - \vtheta_{t-1}}^2} + 3\nu_h^2 \E{\twonorm{\vtheta_t - \vtheta_{t-1}}^2 } \\
	&= \left( 6L^2 + 3\nu_h^2 \right)\E{\twonorm{\vtheta_t - \vtheta_{t-1}}^2},
\end{align*}	
where the last inequality is due to Lemma \ref{lmm:smooth} and the fact $\E{\twonorm{ \left( \mH( \hat{\tau}_{t}; \hat{\vtheta}_t ) - \nabla^2 J_\rho^{H}(\hat{\vtheta}_t) \right) (\vtheta_t - \vtheta_{t-1})}^2} 
 \leq \nu_h^2 \E{\twonorm{\vtheta_t - \vtheta_{t-1}}^2}$. This fact can be proved as follows:
\begin{align*}
&\E{\twonorm{ \left( \mH( \hat{\tau}_{t}; \hat{\vtheta}_t ) - \nabla^2 J_\rho^{H}(\hat{\vtheta}_t) \right) (\vtheta_t - \vtheta_{t-1})}^2} \\
&= \E{ \E[\hat{\tau}_t]{\twonorm{ \left( \mH( \hat{\tau}_{t}; \hat{\vtheta}_t ) - \nabla^2 J_\rho^{H}(\hat{\vtheta}_t) \right) (\vtheta_t - \vtheta_{t-1})}^2}} \\
&=\E{ \E[\hat{\tau}_t]{\twonorm{  \mH( \hat{\tau}_{t}; \hat{\vtheta}_t) (\vtheta_t - \vtheta_{t-1})}^2} - \twonorm{\nabla^2 J_\rho^{H}(\hat{\vtheta}_t)(\vtheta_t - \vtheta_{t-1})}^2}\\
&\leq \E{ \E[\hat{\tau}_t]{\twonorm{  \mH( \hat{\tau}_{t}; \hat{\vtheta}_t) (\vtheta_t - \vtheta_{t-1})}^2} }\\
&\leq \nu_h^2 \E{\twonorm{\vtheta_t - \vtheta_{t-1}}^2},
\end{align*}
where the last line is due to Lemma \ref{lmm:variance}. 
A direct computation yields that
\begin{align*}
	\vu_t - \nabla J_\rho^H(\vtheta_t) &= (1-\beta_t) \left( \vu_{t-1} +  \mH( \hat{\tau}_{t}; \hat{\vtheta}_t )(\vtheta_t - \vtheta_{t-1})\right) +  \vg(\tau_{t}; \vtheta_t) - \nabla J_\rho^H(\vtheta_t)\\
	&=(1-\beta_t) \left( \vu_{t-1} - \nabla J_\rho^H(\vtheta_{t-1}) \right)  + (1-\beta_t) \left(  \nabla J_\rho^H(\vtheta_{t-1}) -  \nabla J_\rho^H(\vtheta_{t}) +  \mH( \hat{\tau}_{t}; \hat{\vtheta}_t )(\vtheta_t - \vtheta_{t-1})\right)\\
	&\quad + \beta_t \left( \vg(\tau_{t}; \vtheta_t) - \nabla J_\rho^H(\vtheta_t) \right)\\
	&=(1-\beta_t) \left( \vu_{t-1} - \nabla J_\rho^H(\vtheta_{t-1}) \right)  + (1-\beta_t) \calH_t + \beta_t \calG_t.
\end{align*}
Applying the above decomposition, we have
\begin{align*}
	V_t &=\E{\twonorm{\vu_t - \nabla J_\rho^H(\vtheta_t) }^2}\\
	&= \E{\twonorm{(1-\beta_t) \left( \vu_{t-1} - \nabla J_\rho^H(\vtheta_{t-1}) \right)  + (1-\beta_t) \calH_t + \beta_t \calG_t }^2}\\
	&\stackrel{(a)}{=}(1-\beta_t)^2 \E{\twonorm{\vu_{t-1} - \nabla J_\rho^H(\vtheta_{t-1}) }^2} + \E{\twonorm{(1-\beta_t) \calH_t + \beta_t \calG_t}^2}\\
	&\leq (1-\beta_t) \E{\twonorm{\vu_{t-1} - \nabla J_\rho^H(\vtheta_{t-1}) }^2} + 2(1-\beta_t)^2\E{\twonorm{ \calH_t}^2} + 2\beta_t^2\E{\twonorm{ \calG_t}^2}\\
	&\stackrel{(b)}{\leq }(1-\beta_t) \E{\twonorm{\vu_{t-1} - \nabla J_\rho^H(\vtheta_{t-1}) }^2} + 2\E{\twonorm{ \calH_t}^2} + 2\beta_t^2\E{\twonorm{ \calG_t}^2}\\
	&\leq (1-\beta_t) V_{t-1} + 2\left(6L^2 + 3\nu_h^2\right) \E{\twonorm{\vtheta_t - \vtheta_{t-1}}^2} + 2\beta_t^2 \nu_g^2,
\end{align*}
where step (a) is due to that $\E{\la\vu_{t-1} - \nabla J_\rho^H(\vtheta_{t-1}), (1-\beta_t) \calH_t + \beta_t \calG_t \ra}=0$ and step (b) follows from the assumption $\beta\leq 1$. Thus we complete the proof.

\subsection{Proof of Lemma \ref{lemma 3}}
Since $\alpha_0= \sqrt{\frac{\mu_F^2}{\kappa\tau_0(12L^2 + 6\nu_h^2)}} \leq \frac{\mu_F}{2L} $, Lemma \ref{lemma ascent 2} holds.  Recall that $\Lambda_t = \Delta_t + \lambda_{t-1} V_t$. Suppose $\Delta_t \geq \frac{2\sqrt{\varepsilon_{\bias}}}{1-\gamma}$. A direct computation yields that
	\begin{align*}
		\Lambda_{t+1} &=  \Delta_{t+1}+ \lambda_{t} V_{t+1} \\
		& \leq \Delta_t - \frac{c_1\alpha_t}{4\kappa} \Delta_t^2  + \frac{4\alpha_tM_g}{\mu_F^2}  V_t  - \frac{\mu_F}{4\alpha_t} R_t +  \frac{ \alpha_t \varepsilon_{\bias}}{2\kappa (1-\gamma)^2} + \frac{4\alpha_tM_g}{\mu_F^2} G_g^2 \gamma^{2H}\\
		&\quad + \lambda_{t} \left((1-\beta_{t+1})V_{t} + \left(12L^2 + 6\nu_h^2\right) R_{t} + 2\beta_{t+1}^2 \nu_g^2\right) - \lambda_{t-1} V_t +  \lambda_{t-1} V_t\\
		&= \Lambda_t - \frac{c_1 \alpha_t}{4\kappa}\Delta_t^2  + \left( \frac{4\alpha_tM_g}{\mu_F^2} + \lambda_t - \lambda_{t-1} - \lambda_t \beta_{t+1} \right)V_t  - \left(\frac{\mu_F}{4\alpha_t} - \lambda_t \left(12L^2 + 6\nu_h^2\right)  \right)R_t  \\
		&\quad + 2\lambda_t \beta_{t+1}^2 \nu_g^2 + \frac{\alpha_t}{2\kappa} \frac{  \varepsilon_{\bias}}{(1-\gamma)^2} + \frac{4\alpha_tM_g}{\mu_F^2} G_g^2 \gamma^{2H}\\
		&\stackrel{(a)}{\leq}\Lambda_t - \frac{c_1 \alpha_t}{4\kappa}\Delta_t^2  -\frac{\alpha_t  M_g}{2\mu_F^2} V_t  + \frac{\alpha_t}{2\kappa} \frac{  \varepsilon_{\bias}}{(1-\gamma)^2} + \frac{4\alpha_tM_g}{\mu_F^2} G_g^2 \gamma^{2H}+ 2\lambda_t \beta_{t+1}^2 \nu_g^2\\
		&\stackrel{(b)}{\leq }\Lambda_t - \frac{c_1 \alpha_t}{4\kappa}\Delta_t^2  -\frac{\alpha_t  M_g}{2\mu_F^2} V_t  + \frac{c_1 \alpha_t}{8\kappa}\Delta_t^2 + \frac{4\alpha_tM_g}{\mu_F^2} G_g^2 \gamma^{2H}+ 2\lambda_t \beta_{t+1}^2 \nu_g^2\\
		&=\Lambda_t - \frac{c_1 \alpha_t}{8\kappa}\Delta_t^2  -\frac{\alpha_t  M_g}{2\mu_F^2} V_t  + \frac{4\alpha_tM_g}{\mu_F^2} G_g^2 \gamma^{2H}+ 2\lambda_t \beta_{t+1}^2 \nu_g^2, \numberthis \label{eq T2}
	\end{align*}
	where step (b) follows from the assumption $\varepsilon_{\bias} \leq (1-\gamma)^2 c_1 \Delta^2_t/4$. Since $\alpha_t = \alpha_0 \beta_t^{1/2}, \lambda_t = \lambda_0 \beta_t^{-1/2}, \lambda_0 =  \frac{\kappa \tau_0\alpha_0}{4\mu_F}$ and $\alpha_0 =\sqrt{\frac{\mu_F^2}{\kappa \tau_0(12L^2 + 6\nu_h^2)}}$, we have
	\begin{align*}
		\frac{\mu_F}{4\alpha_t} - \lambda_t (12L^2 + 6\nu_h^2) &=\frac{\mu_F}{4 \alpha_0 \beta_t^{1/2}}  - \frac{\lambda_0 (12L^2 + 6\nu_h^2)}{\beta_t^{1/2}}  =  \frac{\mu_F}{4 \alpha_0 \beta_t^{1/2}}  - \frac{ (12L^2 + 6\nu_h^2)}{\beta_t^{1/2} } \cdot  \frac{ \kappa\tau_0  \alpha_0}{4\mu_F} \\
  &= \frac{\mu_F }{4 \alpha_0\beta_t^{1/2}} \left( 1- \frac{\kappa \tau_0 \alpha_0^2(12L^2 + 6\nu_h^2)}{\mu_F^2}\right) =0.
	\end{align*}
	Additionally, a direct computation leads to that
	\begin{align*}
		\lambda_t - \lambda_{t-1} &= \frac{\lambda_0}{\beta_t^{1/2}} -  \frac{\lambda_0}{\beta_{t-1}^{1/2}}  =\frac{ \lambda_0}{\sqrt{\tau_0}} \left(\sqrt{t+\tau_0} - \sqrt{t-1+\tau_0}\right) \\
		&= \frac{\lambda_0}{\sqrt{\tau_0}}\left(\frac{1}{\sqrt{t+\tau_0} + \sqrt{t-1+\tau_0}}\right) \leq \frac{\lambda_0}{\tau_0} \frac{\sqrt{\tau_0}}{\sqrt{t+\tau_0}  }=\frac{\lambda_0}{\tau_0}\beta_t^{1/2},\\
		\beta_{t+1} &=\frac{\tau_0}{t+1+\tau_0} = \frac{t+\tau_0}{t+1+\tau_0}\frac{\tau_0}{t+\tau_0} \geq \frac{\tau_0}{\tau_0+1} \beta_t(\text{ for }t\geq 0) \geq \frac{19}{\tau_0} \beta_t (\text{ for }\tau_0 \geq 20), \\
		-\lambda_t \beta_{t+1} &\leq -\frac{19}{\tau_0} \lambda_t\beta_t =-\frac{19\lambda_0\beta_t^{1/2}}{\tau_0},\\
		\frac{4M_g \alpha_t}{\mu_F^2}  + \lambda_t - \lambda_{t-1} - \lambda_t \beta_{t+1} &\leq  \frac{4M_g \alpha_t}{\mu_F^2}  + \frac{\lambda_0}{\tau_0}\beta_t^{1/2}-\frac{19\lambda_0\beta_t^{1/2}}{\tau_0} = \frac{4M_g \alpha_t}{\mu_F^2}  -\frac{18\lambda_0\beta_t^{1/2}}{\tau_0} \\
		&= \frac{4M_g \alpha_t}{\mu_F^2}  - \frac{18\tau_0 M_g \alpha_0}{4\mu_F^2} \cdot \frac{\beta_t^{1/2}}{\tau_0} \\ 
		&=  -\frac{M_g}{2\mu_F^2} \alpha_t,
	\end{align*}
	where we have used the fact that $\lambda_0 = \frac{\tau_0 M_g \alpha_0}{4\mu_F^2}$. Thus step (a) holds. Recall that $\Lambda_t = \Delta_t + \lambda_{t-1} V_t$, one has
	\begin{align*}
		T:&= - \frac{c_1 \alpha_t}{8\kappa}\Delta_t^2  -\frac{\alpha_t  M_g}{2\mu_F^2} V_t \\
		&=- \frac{c_1 \alpha_t}{8\kappa}\left( \Lambda_t - \lambda_{t-1} V_t\right)^2  -\frac{\alpha_t  M_g}{2\mu_F^2} V_t \\
		&=\frac{\alpha_t c_1}{8\kappa} \left( - \left( \Lambda_t - \lambda_{t-1} V_t\right)^2  -\frac{4\kappa^2}{\mu_Fc_1} V_t \right)\\
		&\leq -\frac{\alpha_t\kappa}{2\mu_F \lambda_{t-1}} \Lambda_t + \frac{\alpha_t\kappa^3}{ 2c_1 \mu_F^2 \lambda_{t-1}^2}, \numberthis \label{eq T}
	\end{align*}
	where the first inequality is due to $-(a_1 x - b)^2 - a_2 x \leq -\frac{a_2}{a_1} b + \frac{a_2^2}{4a_1^2}$ with $a_1=\lambda_{t-1}, a_2=\frac{4\kappa^2}{\mu_Fc_1}  $ and $b=\Lambda_t$. Plugging \eqref{eq T} into \eqref{eq T2} yields that
	\begin{align*}
		\Lambda_{t+1} &\leq \Lambda_t - \frac{c_1 \alpha_t}{8\kappa}\Delta_t^2  -\frac{\alpha_t  M_g}{2\mu_F^2} V_t  + \frac{4\alpha_tM_g}{\mu_F^2} G_g^2 \gamma^{2H}+ 2\lambda_t \beta_{t+1}^2 \nu_g^2\\
		&\leq \Lambda_t -\frac{\alpha_t\kappa}{2\mu_F \lambda_{t-1}} \Lambda_t + \frac{\alpha_t\kappa^3}{ 2c_1 \mu_F^2 \lambda_{t-1}^2}+ \frac{4\alpha_tM_g}{\mu_F^2} G_g^2 \gamma^{2H}+ 2\lambda_t \beta_{t+1}^2 \nu_g^2\\
		&\leq \Lambda_t -\frac{2\beta_t}{\tau_0} \Lambda_t +  \frac{16\kappa \beta_t^{3/2}}{c_1 \alpha_0\tau_0^2}  + \frac{\tau_0 \kappa\alpha_0 \nu_g^2}{2\mu_F} \beta^{3/2}_t + \frac{4\alpha_0 M_gG_g^2 \gamma^{2H}  }{\mu_F^2}  \beta_t^{1/2}\\
		&= \left(1 -\frac{2\beta_t}{\tau_0} \right) \Lambda_t + \left( \frac{16\kappa }{c_1 \alpha_0\tau_0^2}  + \frac{\tau_0 \kappa\alpha_0 \nu_g^2}{2\mu_F} \right)\beta^{3/2}_t + \frac{4\alpha_0 M_gG_g^2 \gamma^{2H}  }{\mu_F^2}  \beta_t^{1/2},
	\end{align*}
	where the third line is due to the following relationships:
	\begin{align*}
		\beta_t &\leq \beta_{t-1} = \frac{\tau_0}{t-1+\tau_0} = \frac{t + \tau_0}{t-1+\tau_0}  \frac{\tau_0}{t+\tau_0} \leq 2\beta_t \quad (\text{ for }t,\tau_0\geq 1),\\
		-\frac{\alpha_t\kappa}{2\mu_F \lambda_{t-1}} & = -\frac{\kappa}{2\mu_F} \cdot \frac{\alpha_0 \beta_t^{1/2}}{\lambda_0 \beta_{t-1}^{-1/2}} \leq -\frac{\kappa}{2\mu_F} \cdot \frac{\alpha_0}{\lambda_0}  \beta_t= -\frac{\kappa}{ 2\mu_F} \cdot \frac{4\mu_F}{\tau_0 \kappa} \beta_t = -\frac{2\beta_t}{\tau_0},\\
		\frac{\alpha_t\kappa^3}{ 2c_1 \mu_F^2 \lambda_{t-1}^2} &= \frac{\kappa^3}{ 2c_1 \mu_F^2} \cdot \frac{\alpha_0 \beta_t^{1/2}}{ \lambda_0^2 \beta_{t-1}^{-1}} = \frac{\kappa^3}{ 2c_1 \mu_F^2} \cdot \frac{\alpha_0 \beta_t^{1/2}  \beta_{t-1}}{ \lambda_0^2} \\
		& \leq \frac{\kappa^3}{ 2c_1 \mu_F^2} \cdot \frac{2\alpha_0 \beta_t^{3/2} }{ \lambda_0^2}=\frac{\kappa^3}{ 2c_1 \mu_F^2} \cdot  2\alpha_0 \beta_t^{3/2} \left(\frac{4\mu_F}{\tau_0 \kappa \alpha_0} \right)^2=\frac{16\kappa \beta_t^{3/2}}{c_1\alpha_0 \tau_0^2},\\
		2\lambda_t \beta_{t+1}^2  &=2\lambda_0 \beta_t^{-1/2} \beta_{t+1}^2 \leq 2\lambda_0 \beta_t^{-1/2} \beta_{t}^2 = 2\lambda_0 \beta_t^{3/2} = \frac{\tau_0 \kappa\alpha_0}{2\mu_F} \beta^{3/2}_t,
	\end{align*}
	where we have used the fact $\lambda_0 = \frac{\tau_0\kappa \alpha_0}{4\mu_F}$.

\bibliographystyle{plain}
\bibliography{refs}
\end{document}